\documentclass{article}

\usepackage[preprint]{neurips_2025}

\usepackage{natbib}
\usepackage{amssymb}
\usepackage{pifont}
\usepackage{tabularx} % 自动换行的表格列
\usepackage{booktabs} % 更漂亮的表格线
\usepackage{subcaption}  % 加在 preamble 中

\usepackage[utf8]{inputenc} % allow utf-8 input
\usepackage[T1]{fontenc}    % use 8-bit T1 fonts
\usepackage{hyperref}       % hyperlinks
\usepackage{url}            % simple URL typesetting
\usepackage{booktabs}       % professional-quality tables
\usepackage{amsfonts}       % blackboard math symbols
\usepackage{nicefrac}       % compact symbols for 1/2, etc.
\usepackage{microtype}      % microtypography
\usepackage{xcolor}         % colors
\usepackage{enumitem}
\usepackage{graphicx}
\usepackage{amsmath}
\usepackage{multirow}
\newcommand{\xmark}{\ding{55}} % 叉号

\title{StyleDrive: Towards Driving-Style Aware Benchmarking of End-To-End Autonomous Driving}
\author{%
  Ruiyang Hao$^{1,2}$, Bowen Jing$^{3}$, Haibao Yu$^{1,4}$, Zaiqing Nie$^{1,}$\thanks{Corresponding to zaiqing@air.tsinghua.edu.cn.} \\
  $^{1}$ AIR, Tsinghua University\;  $^{2}$ King's College London \\
  $^{3}$ The University of Manchester\; $^{4}$ The University of Hong Kong\\
  \normalsize{
    ~\url{https://styledrive.github.io/}
    }
}

\begin{document}

\maketitle

\begin{abstract}
Personalization, while extensively studied in conventional autonomous driving pipelines, has been largely overlooked in the context of end-to-end autonomous driving (E2EAD), despite its critical role in fostering user trust, safety perception, and real-world adoption. 
A primary bottleneck is the absence of large-scale real-world datasets that systematically capture driving preferences, severely limiting the development and evaluation of personalized E2EAD models. 
In this work, we introduce the first large-scale real-world dataset explicitly curated for personalized E2EAD, integrating comprehensive scene topology with rich dynamic context derived from agent dynamics and semantics inferred via a fine-tuned vision-language model (VLM). 
We propose a hybrid annotation pipeline that combines behavioral analysis, rule-and-distribution-based heuristics, and subjective semantic modeling guided by VLM reasoning, with final refinement through human-in-the-loop verification. 
Building upon this dataset, we introduce the first standardized benchmark for systematically evaluating personalized E2EAD models. 
Empirical evaluations on state-of-the-art architectures demonstrate that incorporating personalized driving preferences significantly improves behavioral alignment with human demonstrations.
\end{abstract}

\section{Introduction}

As autonomous driving (AD) technologies continue to mature, one of the key factors for translating these systems into real-world products is personalization. Tailoring vehicle behavior to individual user preferences is essential for enhancing user experience, building trust, and fostering long-term
adoption~\cite{hasenjager2017personalization}. Traditional modular systems have extensively explored personalized strategies, enabling adaptations to specific driving preferences. However, these methods often rely on isolated, scenario-specific adaptations~\cite{tian2022personalized} or unrealistic human-in-the-loop (HITL) simulation~\cite{ke2024d2e}, which severely limit their generalization to dynamic, real-world environments. Furthermore, the fragmented nature of modular systems~\cite{zhu2018personalized, cui2024personalized, kou2025padriver} hinders scalability to massive real-world data. Due to these limitations, personalization remains largely underexplored in end-to-end autonomous driving (E2EAD), where perception, planning, and control are integrated in an unified architecture~\cite{codevilla2018end}. This gap presents a significant barrier to realizing human-centric AD at scale.

\begin{figure*}
  \centering
  \includegraphics[width=0.9\linewidth]{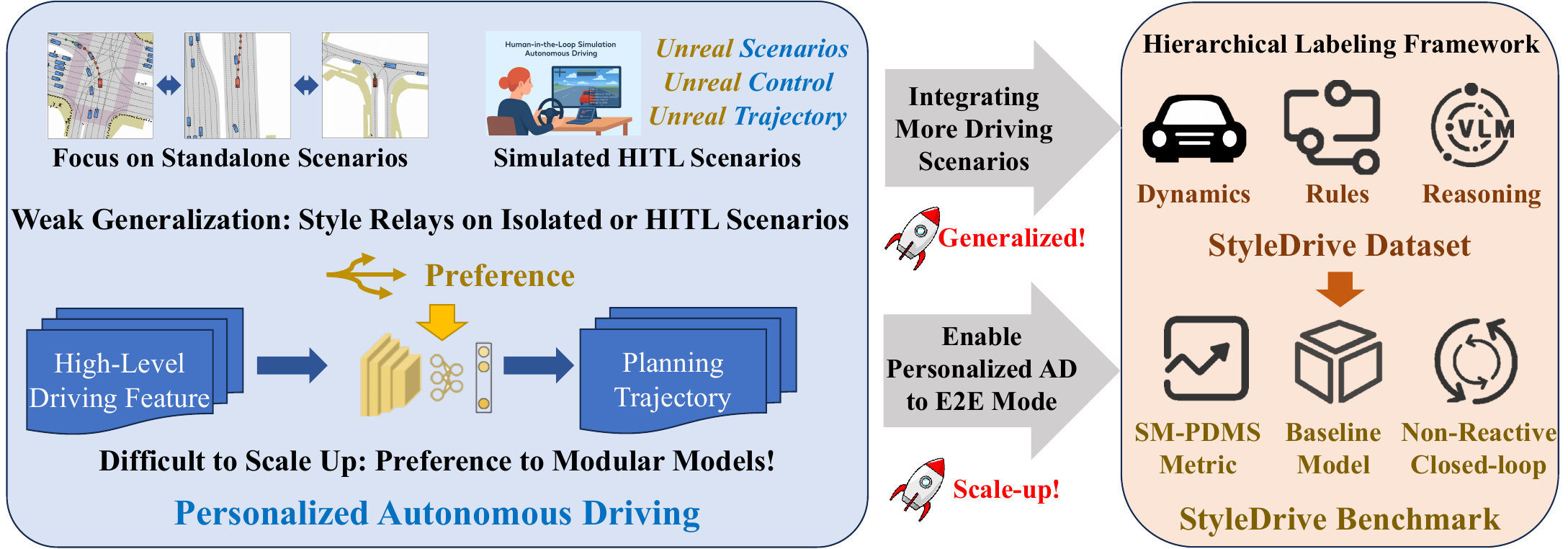}
      \caption{Motivation and Overview of StyleDrive.}
  \label{fig:overview}
\end{figure*}

This gap is particularly pressing. As E2EAD systems become increasingly capable and are deployed across diverse, open-world driving environments, aligning vehicle behavior with user preferences becomes essential for enhancing comfort, perceived safety, and long-term user acceptance~\cite{speidel2019towards, aledhari2023motion}. Yet integrating personalization into the end-to-end paradigm introduces unique challenges. Unlike modular systems operating within narrowly defined scenarios, E2EAD requires preference generalization across a broad spectrum of complex, real-world scenarios. Meeting this demand necessitates large-scale datasets that not only cover diverse traffic conditions but also provide well-crafted driving style annotations.

To fill this critical gap, we introduce \textbf{the first large-scale real-world dataset purposefully designed for personalized E2EAD research}. Our dataset captures both objective behavioral patterns and subjective driving preferences across a wide range of driving scenarios. Static environmental features are firstly extracted from real-world road topologies, while dynamic context cues are inferred using a fine-tuned visual language model (VLM), enabling rich semantic scene understanding. Based on these features, we generate objective annotations through behavior distribution analysis and rule\&distribution-based heuristics. To incorporate subjectivity, we further employ the VLM to label preferences considering static and dynamic scene semantics. The pipeline ends with a label fusion mechanism to ensure quality and consistency. Building upon this dataset, we establish \textbf{the first benchmark for evaluating personalized E2EAD models}. We conduct extensive experiments on multiple state-of-the-art architectures, with and without preference conditioning, and show that incorporating personalization significantly improves alignment with human-like driving behavior.

The motivation and overview of our work are illustrated in Fig.~\ref{fig:overview}, aiming to bridge the gap between the growing momentum in E2EAD and the longstanding need for personalization in AD. In summary, our key contributions are:
\begin{itemize}[leftmargin=1em, topsep=0pt, itemsep=0.3em]
    \item \textbf{A novel large-scale real-world dataset} for personalized end-to-end autonomous driving, annotated with both objective behaviors and subjective driving style preferences across a wide range of traffic scenarios.
    \item \textbf{A multi-stage annotation pipeline} that integrates behavior feature analysis, rule\&distribution-based heuristics, VLM reasoning, and human-in-the-loop validation to generate high-quality well-crafted preference labels.
    \item \textbf{The first benchmark for personalized E2EAD}, which enables standardized, quantitative comparison of preference-conditioned behavior across four different SOTA model architectures.
    \item \textbf{Comprehensive empirical results} demonstrating that incorporating user driving preferences significantly improves alignment with human-like behavior, underscoring the value of personalization in E2EAD.
\end{itemize}

\begin{table*}[ht]
\centering
\caption{
Comparison of driving datasets by data source, scenario coverage, evaluation protocol, end-to-end (E2E) support, and driving style annotations. 
\textbf{StyleDrive} is the only dataset that combines real-world data, diverse scenarios, semi-closed-loop (Semi-CL) evaluation, E2E learning capability, and structured driving style labels for personalized autonomous driving.
\\
\footnotesize{\textit{Abbreviations:} 
OL = Open-Loop, 
CL = Closed-Loop, 
Semi-CL = Semi-Closed-Loop, 
HITL = Human-in-the-Loop, 
E2E = End-to-End driving model support,
Style = With Style or Preference Annotation}
}
\setlength{\tabcolsep}{1mm}
\begin{tabular}{lcccccc}
\hline
\hline
\textbf{Dataset} & \textbf{Reality} & \textbf{Scenarios} & \textbf{CL/OL} & \textbf{E2E} & \textbf{Style} \\
\hline
nuScenes~\cite{caesar2020nuscenes} & Real   & City           & OL           & \checkmark  & \xmark \\
OpenScene~\cite{peng2023openscene}   & Real     & City\&Rural         & OL            & \checkmark  & \xmark \\
Longest6~\cite{chitta2022transfuser} & Sim     & City         & CL            & \checkmark  & \xmark \\
CARLA~\cite{carla_leaderboard_2024}  & Sim     & City         & CL            & \checkmark  & \xmark \\
MetaDrive~\cite{li2022metadrive}  & Sim     & City         & CL            & \checkmark  & \xmark \\
Bench2Drive~\cite{jia2024bench2drive}  & Sim     & City         & CL            & \checkmark  & \xmark \\
NAVSIM~\cite{dauner2024navsim}  & Real     & City\&Rural         & Semi-CL            & \checkmark  & \xmark \\ 
\hline
HITL-RampMerging~\cite{li2023personalized} & HITL   & Ramp-Merge      & OL            & \xmark      & \checkmark \\
HITL-CarFollowing~\cite{zhao2022personalized} & HITL   & Car-Following       & OL           & \xmark      & \checkmark \\
HITL-LaneChange~\cite{liao2023driver} & HITL & Lane-Change & OL          & \xmark      & \checkmark \\
HITL-MultiScene~\cite{ke2024d2e} & HITL & City & OL          & \xmark      & \checkmark \\
UAH~\cite{romera2016need} & Real   & City\&highway      & OL            & \xmark      & \checkmark \\
Brain4Cars~\cite{jain2016brain4cars} & Real & Lane-Change\&Merge & OL          & \xmark      & \checkmark \\
PDB~\cite{wei2025pdb} & Real   & City       & OL           & \xmark      & \checkmark \\
\hline
\textbf{StyleDrive (Ours)}      & Real      & City\&Rural       & Semi-CL     & \checkmark  & \checkmark \\
\hline
\hline
\end{tabular}
\label{tab:style-dataset-comparison}
\end{table*}

\section{Related Work}
\paragraph{Personalized Autonomous Driving}
Personalization has long been recognized as a key factor to enhance user comfort, trust, and acceptance of AD~\cite{liao2024review}. In conventional modular pipelines, personalized strategies have been extensively explored, where the system can adapt to users' preferences in stand-alone driving scenarios, such as car following, ramp merge, and lane-change behaviors~\cite{zhao2022personalized, li2023personalized, liao2023driver}. Other early attempts personalize models in unrealistic human-in-the-loop (HITL) simulation~\cite{ke2024d2e}. However, these approaches often suffer from poor generalization in dynamic, real-world environments.

Recent advances in large language models (LLMs) have introduced new possibilities for personalization in AD, enabling the encoding of user intent and preferences to influence downstream decision-making~\cite{cui2024personalized, xu2024drivegpt4, kou2025padriver}. Despite their potential, these methods remain constrained by modular frameworks and often depend on handcrafted features. MAVERIC~\cite{schrum2024maveric} represents a notable early attempt to enable end-to-end personalized driving. Nevertheless, in the absence of standardized datasets and benchmarks, these methods face challenges in reproducibility and scalability. Despite these advancements, existing work seldom addresses how to merge user preferences in an E2EAD manner. A critical bottleneck is the lack of large-scale, real-world datasets with well-crafted annotations of personalized driving styles.

\paragraph{End-to-End Autonomous Driving}

End-to-end autonomous driving represents a promising paradigm that maps raw sensor inputs directly to driving actions or planned trajectories. Early work~\cite{ codevilla2018end} demonstrated the feasibility of E2EAD. Over the past decade, research has advanced toward more sophisticated architectures incorporating temporal reasoning, multimodal fusion, and Transformer-based methods~\cite{prakash2021multi, jia2023think, shao2023safety}.

Despite this progress, most E2EAD models are optimized for average-level objectives, lacking the capacity to adapt to user-specific preferences. While recent architectures improve generalization and interpretability~\cite{li2024enhancing, zheng2024genad, jia2025drivetransformer}, they remain limited in capturing individualized driving styles. Moreover, the absence of dedicated datasets and standardized benchmarks for evaluating personalization continues to hinder systematic advancement in this area. To address these limitations, we propose a framework for learning style-conditioned E2EAD models and establish a benchmark to evaluate their alignment with human driving behavior.
% Our work represents the first systematic attempt to align end-to-end driving behavior with real-world personalized objectives, paving the way for more adaptive and user-aligned autonomous vehicles.

\paragraph{Benchmarking Autonomous Driving}
Benchmarking is fundamental to the development and evaluation of AD models. Typical AD datasets and benchmarks are summarized in Tab.~\ref{tab:style-dataset-comparison}. Early E2EAD benchmarks~\cite{caesar2020nuscenes,peng2023openscene} adopt non-interactive open-loop settings with real-world data, but fail to capture behavioral feedback. Recent works have introduced closed-loop benchmarks in simulation, such as Longest6~\cite{chitta2022transfuser},  MetaDrive~\cite{li2022metadrive}, CARLA Series~\cite{carla_leaderboard_2024}, which allow online policy evaluation but rely entirely on simulated scenarios. Semi-simulated platforms like NAVSIM~\cite{dauner2024navsim} strike a balance by simulating behavior feedback based on real-world scenes. However, existing benchmarks remain focused on task performance and ignore user preferences, cannot evaluate personalized driving behavior.

Efforts to benchmark personalized AD have led to two main dataset categories: human-in-the-loop (HITL) simulations and real-world driving logs. 1) HITL datasets are typically collected in controlled scenarios such as ramp merging~\cite{li2023personalized}, car following~\cite{zhao2022personalized}, lane changing~\cite{liao2023driver}, and multiple scenarios~\cite{ke2024d2e}. While these setups enable preference modeling, they often rely on simplified simulation engines and curated environments, introducing a domain gap between real-world conditions and unrealistic scenarios. 2) Real-world datasets such as UAH-DriveSet~\cite{romera2016need}, Brain4Cars~\cite{jain2016brain4cars}, and PDB~\cite{wei2025pdb} provide behavioral annotations related to diving intent. These datasets have contributed substantially to understanding driver behavior and intent modeling. However, they are typically constrained to narrow driving contexts, such as highways or intersections. More importantly, they are not structured for end-to-end learning: most only provide high-level behavior labels without continuous perception-to-control supervision, and none include standardized evaluation protocols for comparing personalized driving policies.

To bridge the gap between E2EAD and personalized AD, we introduce the first benchmark tailored for personalized E2EAD. Built on real-world data, our benchmark supports both supervised learning and semi-closed-loop evaluation of preference-conditioned policies, enabling reproducible evaluation of personalized E2E across diverse scenarios.

\section{StyleDrive Dataset}

The StyleDrive dataset is sampled and constructed from the large-scale AD dataset OpenScene~\cite{peng2023openscene}, and comprises nearly 30k driving scenarios annotated with style preferences. OpenScene features trajectories collected from over 16 drivers across diverse urban and suburban environments in both Singapore and the USA. This diversity offers a rich foundation for labeling user-specific tendencies in various scenarios (also illustrated in Fig.~\ref{fig:figure_demo}). Beyond the original OpenScene data, we introduce a unified framework for modeling and annotating personalized driving preferences.

\begin{figure*}[ht]
  \centering
  \includegraphics[width=0.90\linewidth]{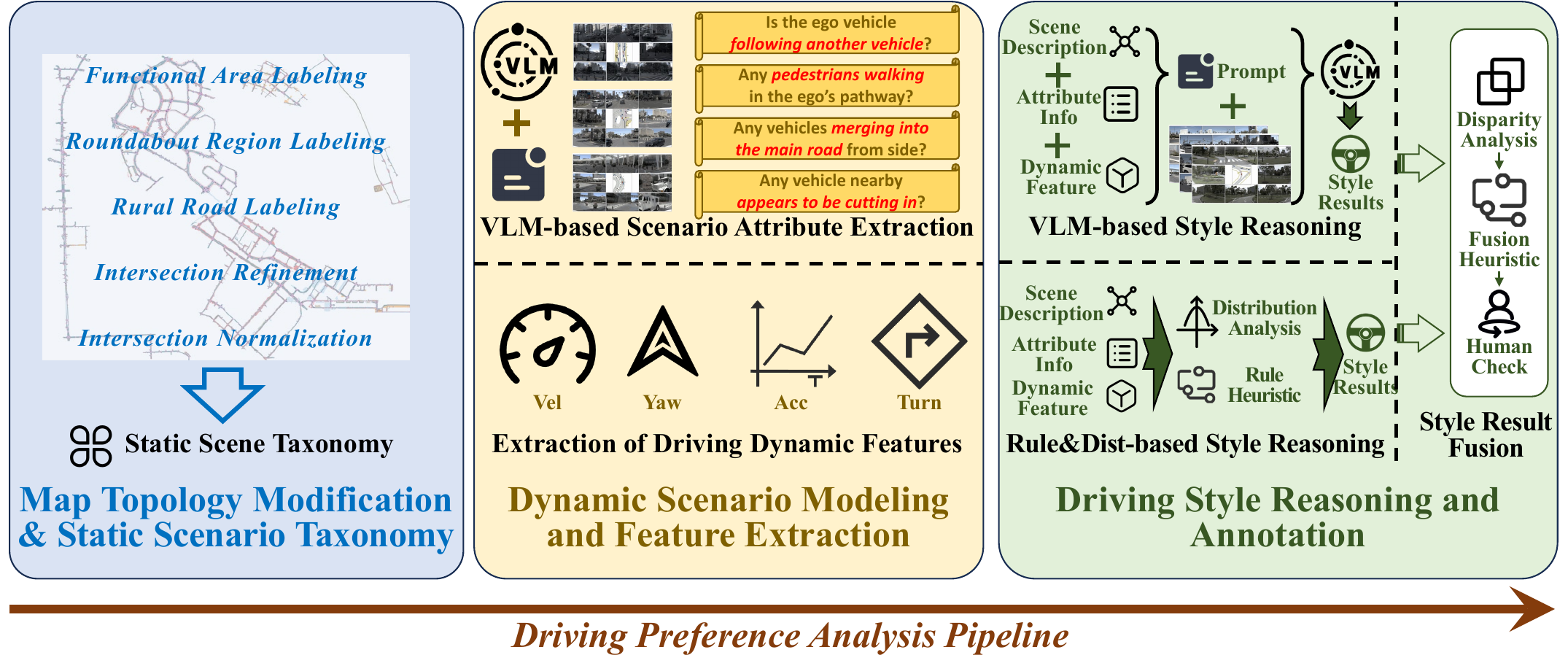}
  \caption{Framework for Modeling and Annotation of Driving Preference.}
  \label{fig:anno_framework}
\end{figure*}
\subsection{Framework for Modeling and Annotation of Driving Preference}

\paragraph{Framework Overview.} To enable reliable and interpretable driving style analysis, we construct a hierarchical modeling and annotation framework, as shown in Fig.~\ref{fig:anno_framework}. We extract static environmental features from real-world road topology and dynamic motion features from driving logs, and infer dynamic contextual cues using a fine-tuned visual language model (VLM), enabling consistent and fine-grained scenario construction. From these constructed scenario features, we derive objective preference annotations via behavioral distribution analysis and rule-based heuristics. To capture the inherent subjectivity of driving style, we further leverage the VLM to generate subjective annotations by jointly modeling scene semantics and driver behavior. Final high-quality labels are obtained through a human-in-the-loop verification process that consolidates both perspectives.

\paragraph{Topology-Driven Static Scenario Taxonomy}
To facilitate structured analysis of driving behavior, we first categorize each driving clip into a coarse-grained traffic scenario type based on the topology of high-definition (HD) maps. This categorization provides a physically grounded and semantically interpretable foundation for subsequent scenario modeling. Specifically, we begin by enriching the topological structures based on the original nuPlan HD maps (details in Sect.3 of the Supplementary). We then align each clip data with detailed map elements, such as lane geometries, stop lines, crosswalks, and intersection layouts, to assign it to one of eleven primary traffic scenario types. Each type captures the static constraints and navigational features of the road environment, serving as a structural prior to interpreting the downstream driving behavior. The complete type list is shown in the left part of Fig.~\ref{fig:dataset_stats}.

\paragraph{Dynamic Semantic Refinement via Vision-Language Model}
To enrich the traffic scenario classification with dynamic, context-sensitive information, we apply a fine-tuned vision-language model (VLM) to extract high-level semantic cues from driving videos. Since generic VLMs lack task-specific perception and reasoning capabilities for AD, we adapt the Video-LLaMA3 model~\cite{zhang2025videollama} via lightweight fine-tuning using the LingoQA dataset~\cite{marcu2024lingoqa} - a multimodal dataset for reasoning about road semantics (more details in Sect.5 of the Supplementary).

For each traffic scenario type, we craft targeted prompts related to key driving events. These include questions such as whether a lead car is present in the same lane, whether the ego is merging, or whether pedestrians are visible in crosswalk zones. The model responses are then parsed into structured semantic attributes, such as the presence of lead vehicles, merging behavior, pedestrian involvement, and turning intent. This dynamic semantic layer augments the static topology with interaction-aware, temporally grounded cues, thereby enhancing downstream preference modeling via both rule\&distribution-based and VLM-based reasoning.

\paragraph{Objective Preference Annotation via Rule\&Distribution-based Heuristics}
Building on the enriched scene context, we generate objective driving style annotations using a set of interpretable, physically grounded heuristics informed by motion dynamics and semantic priors. We extract a set of behaviorally relevant physical ego motion features, such as speed, acceleration, yaw rate variation, and proximity to surrounding agents. Based on these ego motion features, structured scene semantics and static scenario type, we analyze the feature distribution and accordingly define scenario-specific rules calibrated from aggregated driving statistics (more details in Sect.4 of the Supplementary). For example:

\begin{itemize}[leftmargin=1em, topsep=0pt, itemsep=1pt]
    \item Low speed and wide safety margins at intersections correspond to conservative tendencies;
    \item Sudden lane changes with low rear headway indicate aggressive behavior;
    \item Moderate-speed following and stable lane keeping fall into the normal category.
\end{itemize}

\noindent Grounded in population-level behavior distributions and further expert validation, these heuristics offer an explicit and robust foundation for objective style labeling.
% These heuristics are grounded on population-level behavioral distributions and refined through expert validation, offering an explicit and robust basis for objective style labeling.

\paragraph{Subjective Preference Annotation via VLM-Based Reasoning}
To complement rule\&distribution-based methods, we leverage the contextual reasoning ability of our fine-tuned Video-LLama3 to generate subjective annotations that reflect human-like interpretations of driving style. Given the driving video, the structured scene semantics and corresponding extracted ego motion features, the model is prompted to answer behavioral questions such as:
\begin{itemize}[leftmargin=1em, topsep=0pt, itemsep=0.5pt]
    \item “Does the ego vehicle appear cautious given the movement of pedestrians?”
    \item “Is the vehicle merging assertively or yielding?”
\end{itemize}
These responses capture high-level intent and interaction cues that extend beyond the expressiveness of fixed heuristics. The multimodal attention of the VLM model enables it to model nuanced behavioral patterns, especially in borderline or semantically complex scenarios.

\paragraph{Multi-Source Fusion and Human-in-the-Loop Verification}
Before finalizing the annotation protocol, we conduct a human-in-the-loop analysis to assess consistency and divergence between rule\&distribution-based and VLM-based labels. Manual inspection reveals the following patterns:

\begin{itemize}[leftmargin=1em, topsep=0pt, itemsep=0.5pt]
    \item Aggressive driving is often consistently detected by both sources. Even when disagreement occurs, each source frequently captures valid signals that the other misses. This complementary behavior suggests that aggressive tendencies are well-suited to permissive merging strategies.
    
    \item In contrast, conservative and normal styles are more easily conflated by VLM, particularly in low-speed or ambiguous interactions. This observation motivates the adoption of stricter criteria for conservative assignments.
\end{itemize}

These observations motivate a risk-aware fusion strategy: 1) If either rule-based or VLM-based reasoning labels a clip as aggressive, we annotate it as aggressive; 2) If both sources agree that the clip reflects conservative behavior, we label it as conservative; 3)  In other cases, the driving style is marked as normal. This strategy emphasizes consistency by being permissive for potential aggressive behavior while remaining strict about conservative assignments. It also ensures that the final label reflects both interpretability and robustness. Besides, final annotations are further verified in edge cases through targeted human review, ensuring label reliability.

\subsection{Dataset Stats, Style Distribution and Visualization}

\begin{figure}
  \centering
  \includegraphics[width=1.0\linewidth]{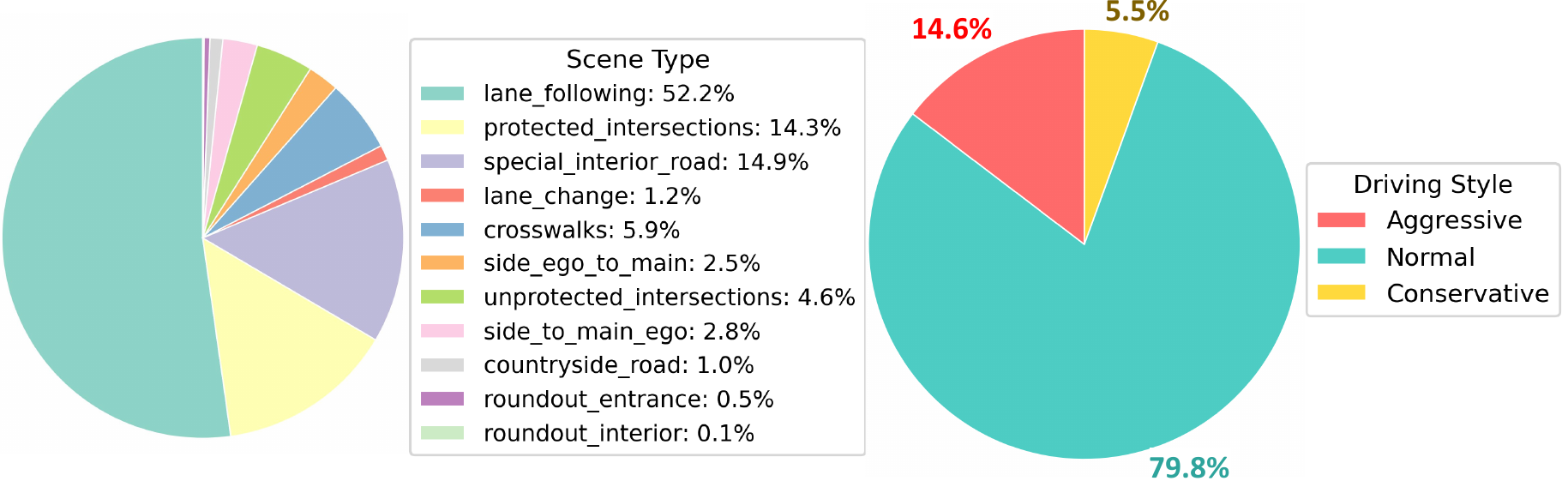}
  \caption{Dataset Statistics and Distribution Analysis.}
  \label{fig:dataset_stats}
\end{figure}

To provide a global overview of the StyleDrive dataset, we present key statistics on both scenario composition and annotated driving preference. As shown in Fig.~\ref{fig:dataset_stats}, we visualize the distribution across static/dynamic scene types and behavioral styles using two complementary pie charts.

\paragraph{Scene Composition.}
The left chart shows the distribution of traffic scenario types.  Lane-follow and intersection scenarios dominate the dataset, reflecting their prevalence in real-world driving. Context-rich types such as merging, lane changes, and pedestrian interactions contribute to diversity.

\paragraph{Driving Style Distribution.}
The right chart illustrates the annotation outcomes for driving preferences. Normal behavior forms the majority, while aggressive and conservative styles are less common but sufficiently represented. This distribution mirrors typical driving patterns and supports balanced learning, especially for edge-case recognition. More dataset stats are provided in Sect.2 of the Supplementary.

\begin{figure}[htbp]
  \centering
  \includegraphics[width=0.86\linewidth]{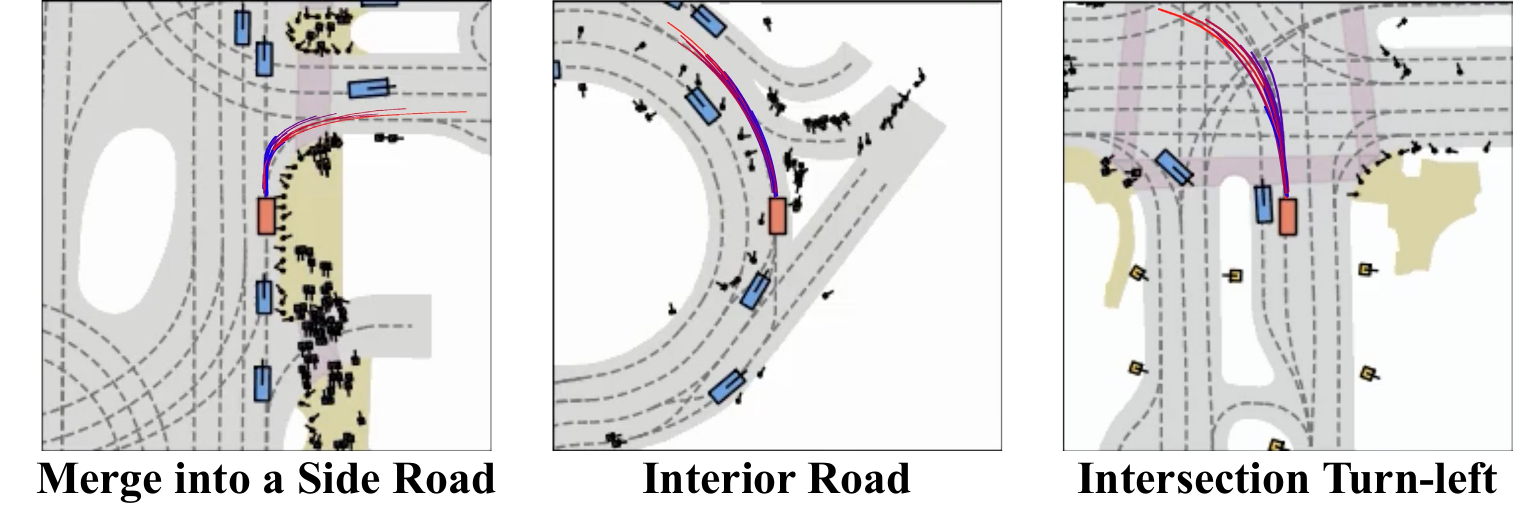}
  \caption{Visualization of driving style distribution in three typical scenarios. Each case is drawn from similar local scenes without pedestrians or leading cars, ensuring style differences arise primarily from drivers' own behavioral preferences. Red trajectories denote aggressive and blue ones denote conservative. More demos are provided in Sect.2 of the Supplementary.}
  \label{fig:figure_demo}
\end{figure}

\paragraph{Driving Style Visualization.}
Fig.~\ref{fig:figure_demo} highlights the distribution of different driving styles. Notably, even under similar driving conditions, drivers exhibit substantial variations in their behavioral preferences.

\section{StyleDrive Benchmark}

To advance the development and evaluation of personalized E2EAD, we introduce the StyleDrive Benchmark, a non-reactive simulation-based evaluation framework for assessing driving preferences and performance in realistic traffic scenarios. This benchmark evaluates whether autonomous agents can generate behavior that aligns with target driving styles while ensuring safety and social compliance. Built upon the rich scenarios and structured annotations of the StyleDrive dataset, the benchmark defines a standardized testbed consisting of four components: simulation environment, the proposed SM-PDMS metric, baseline models, and benchmark results with performance analysis.

\paragraph{Simulation Environment.} 
Our benchmark is built on a lightweight simulator NavSim~\cite{dauner2024navsim}, which emphasizes non-reactive behavior simulation and grounded in real-world traffic scenarios. In our adaptation, the scenarios of StyleDrive are replayed, and style-conditioned E2EAD models learn policies in this playground.

\subsection{Evaluation Metric: Style-Modulated PDMS}

To evaluate whether a model ensures not only safety and feasibility but also alignment with a desired driving style, we propose the Style-Modulated Predictive Driver Model Score (SM-PDMS), an extension of the Predictive Driver Model Score (PDMS) introduced in the NavSim~\cite{dauner2024navsim}. SM-PDMS enhances the original PDMS by incorporating a behavioral alignment component that quantifies the degree to which a policy conforms to specified style preferences. Sub-metrics related to traffic safety and road adherence are retained without modification, as they are considered invariant across driving styles. In contrast, driving style is primarily manifested through motion dynamics, such as following distance, speed, and angular velocity, which in turn affect style-sensitive sub-metrics including comfort (Comf.), ego progress (EP), and time-to-collision (TTC).

Applying the same metric configuration for style-sensitive sub-metrics across different styles is suboptimal for evaluating style-conditioned policies, as users with differing preferences inherently define "better" driving behavior in inconsistent ways. To remedy this, we modulate the style-relevant sub-metrics with annotated styles. Specifically:
\begin{itemize}[leftmargin=1em, topsep=0pt, itemsep=0.5pt]
\item "EP": Ego progress objectives are calibrated to align with target cruising speeds and preferred following distances;
\item "Comf.": Comfort-related thresholds are modulated to reflect varying sensitivities to physical disturbances;
\item "TTC": TTC acceptability ranges are adjusted to capture heterogeneous risk tolerances across styles.
\end{itemize}

\noindent Further details  on these modifications are provided in Sect.1 of the Supplementary. They ensure that SM-PDMS offers a systematic evaluation of both safety and style alignment.

\subsection{Baseline Methods}

To facilitate standardized comparison, we implement four representative style-conditioned baselines covering varied model complexities and paradigms, detailed as follows.

% To establish references for the community, we implement four representative baseline models conditioned on driving style features. These models span a range of model complexities and paradigms, with details remarked as follows:

\begin{itemize}[leftmargin=1em, topsep=0pt, itemsep=1pt, parsep=0pt]
    \item \textbf{AD-MLP-Style}: Extends the AD-MLP baseline~\cite{zhai2023rethinking} with a one-hot driving style vector concatenated to ego states. The combined input is passed through MLPs for style-aware trajectory regression.
    \item \textbf{TransFuser-Style}: Builds on TransFuser~\cite{chitta2022transfuser}, which fuses image and LiDAR features for planning. A one-hot style vector is concatenated with the trajectory query, fused via an MLP to restore dimensionality, and then fed into the trajectory prediction head.
    \item \textbf{DiffusionDrive-Style}: Modifies DiffusionDrive~\cite{liao2024diffusiondrive} by injecting a one-hot style vector into the trajectory head. The vector is concatenated with agent features, fused via a fusion MLP, and followed by a regression MLP. This block runs twice for cascade refinement.
    \item \textbf{WoTE-Style}: Adapts the BEV World model~\cite{li2025end}, which forecasts future BEV states. Driving style is injected into the offset prediction head via concatenation and MLP fusion, following the same strategy as in DiffusionDrive-Style.
\end{itemize}

\noindent Training details are provided in Sect.5 of the Supplementary.

\begin{table*}[t]
  \centering
  \caption{StyleDrive Benchmark Main Results. Style conditioning improves behavioral alignment, with higher SM-PDMS scores across model families. The ablation results (bottom) further confirm the effectiveness and learnability of style conditioning.}
  \setlength{\tabcolsep}{1mm}
    \begin{tabular}{l|cc|>{\centering\arraybackslash}p{1.4cm}
                    >{\centering\arraybackslash}p{1.4cm}
                    >{\centering\arraybackslash}p{1.4cm}|c}
    \hline
    \hline
    \multirow{2}{*}{Models} 
    & \multirow{2}{*}{NC $\uparrow$} 
    & \multirow{2}{*}{DAC $\uparrow$} 
    & \multicolumn{3}{c|}{Style-Modulated Submetrics} 
    & \multirow{2}{*}{SM-PDMS $\uparrow$} \\
    \cline{4-6}
    & & & TTC $\uparrow$ & Comf. $\uparrow$ & EP $\uparrow$ \\
    \hline
    
    AD-MLP \cite{zhai2023rethinking} &  92.63     &  77.68     &  83.83     &   99.75   &  78.01     &    63.72 \\
    
    TransFuser \cite{chitta2022transfuser}    &  96.74      &  88.43      & 91.08      &   99.65     &   84.39     &   78.12       \\

    WoTE \cite{li2025end} &  97.29      &  92.39      & 92.53      &   99.13     &   76.31     &   79.56       \\
    
    DiffusionDrive \cite{liao2024diffusiondrive}    &  96.66       &  91.45      & 90.63      &   99.73      &   80.39     &   79.33      \\ 
    \hline
    
    AD-MLP-Style &  92.38     &  73.23     &  83.14     &   \textbf{99.90}   &  78.55      &    60.02 \\
    
    TransFuser-Style    &  97.23      &  90.36      & 92.61      &   99.73     &   \textbf{84.95}     &   81.09       \\

    WoTE-Style &  \textbf{97.58}      &  \textbf{93.44}      & \textbf{93.70}      &   99.26     &   77.38     &   \textbf{81.38}       \\
    
    DiffusionDrive-Style    &  \textbf{97.81}       &  \textbf{93.45}      & \textbf{92.81}      & \textbf{99.85}      &   \textbf{84.84}     &   \textbf{84.10}      \\ \hline
    - DiffusionDrive-Style-A    &  97.38       &  93.20      & 92.01      & 99.62      &   84.01     &   83.04      \\
    - DiffusionDrive-Style-N    &  97.66       &  93.32      & 92.16      & 99.83      &   84.21     &   83.52      \\
    - DiffusionDrive-Style-C    &  98.23       &  93.59      & 94.98      & 99.87      &   81.36     &   83.90      \\
    \hline
    \hline
    \end{tabular}
  \label{tab:behaviors_analysis}
\end{table*}

\subsection{Main Results Analysis}

\paragraph{Style Conditioning Improves Behavioral Alignment.}
The first two sections of Tab.~\ref{tab:behaviors_analysis} present the quantitative evaluation results of the proposed baseline models on the StyleDrive benchmark. Among the three model families - TransFuser, WoTE, and DiffusionDrive - the style-conditioned variants achieve higher SM-PDMS scores than their vanilla counterparts, clearly demonstrating the benefit of style conditioning. In contrast, the AD-MLP-Style variant slightly underperforms its vanilla version on SM-PDMS. 

Overall, the improvements observed in TransFuser, WoTE, and DiffusionDrive, particularly on style-sensitive metrics like TTC, Comf., and EP, confirm the effectiveness of incorporating style information into planning. The consistent gains of style-conditioned models, especially on Comfort and EP, also provide strong evidence for the validity of our driving style annotations. At an individual level, DiffusionDrive-Style delivers the strongest performance, achieving the highest scores across most metrics. WoTE-Style and TransFuser-Style closely follow, and notably, both outperform the vanilla DiffusionDrive model, further highlighting the benefit of style conditioning. 

The divergent trend in AD-MLP can be attributed to its simplicity: lacking perception, the model cannot effectively leverage style information. Although EP shows a slight improvement, indicating marginally better intent preservation, the drop in DAC ultimately leads to a lower overall score.

\paragraph{Ablation of Fixed Style Conditioning.}
The last section of Tab.~\ref{tab:behaviors_analysis} reports an ablation where the DiffusionDrive-Style model is evaluated by overriding the annotated style and instead enforcing a fixed style condition at test time. For example, DiffusionDrive-Style-A uses the aggressive style condition for all scenes, regardless of ground-truth condition. The results further confirm the effect and learnability of our style conditions: as the conditioned style shifts from aggressive to conservative, NC and DAC show clear improvements. And TTC and Comf. also increase, as SM-PDMS metrics do not penalize overly safe or smooth behaviors. EP peaks with normal style due to its dataset dominance. Besides, fixed-style conditioning underperforms ground-truth style, further validating the reliability of style labels.

\paragraph{Closeness to Human Demonstrations.}
To further assess the behavioral fidelity of style-conditioned models, we conduct an open-loop evaluation using L2 trajectory error against human driving demonstrations. As shown in Tab.~\ref{tab:ole}, style-aware variants consistently exhibit lower prediction errors across all horizons, with DiffusionDrive-Style achieving the best performance. These results highlight that style conditioning not only enhances stylistic expressiveness but also improves direct consistency with human behavior.

\begin{table*}[htbp]
  \centering
  \caption{L2 Open-loop Evaluation Results. Style-conditioned models exhibit lower L2 trajectory error compared to vanilla models, highlighting improved consistency with human driving behavior under preference-aware conditions.}
    \begin{tabular}{l|ccc|c}
    \hline
    \hline
    Models & L2 (2s) $\downarrow$ & L2 (3s) $\downarrow$ & L2 (4s) $\downarrow$ & L2 (Avg) $\downarrow$ \\
    \hline

    WoTE \cite{li2025end} &  0.733     &  1.434    &  2.349     &   1.506 \\
    
    AD-MLP \cite{zhai2023rethinking} &  0.503     &  1.262     &  2.383     &   1.382 \\
    
    TransFuser \cite{chitta2022transfuser}    &  0.431      &  0.963      & 1.701      &   1.032     \\
    
    DiffusionDrive \cite{liao2024diffusiondrive}    &  0.471       &  1.086      & 1.945      &   1.167       \\ \hline

    WoTE-Style &  0.673     &  1.340     &  2.223     &   1.412 \\
    
    AD-MLP-Style &  0.510     &  1.230     &  2.321     &   1.354    \\
    
    TransFuser-Style    &  0.424     &  0.937      & 1.656      &   1.006     \\
    
    DiffusionDrive-Style    &  \textbf{0.417}       &  \textbf{0.940}      & \textbf{1.646}      &   \textbf{1.001}       \\
    \hline
    \hline
    \end{tabular}
  \label{tab:ole}
\end{table*}

\paragraph{Sensitivity Analysis of PDMS vs. SM-PDMS}
Tab.~\ref{tab:sm_pdms_std_range} presents a comparative analysis of the original PDMS and SM-PDMS architectures in terms of standard deviation and range of style-sensitive sub-metrics and final scores in three valid style-conditioned models. SM-PDMS exhibits higher standard deviation and range in final metric scores and style-sensitive sub-metrics such as EP and Comf across models, indicating improved sensitivity to driving style.

\begin{table}[htbp]
  \centering
  \caption{Standard deviation (std.) and range of evaluation metrics across valid style-conditioned models under PDMS and SM-PDMS. SM-PDMS increases sensitivity to style-specific behaviors while maintaining overall consistency, whereas PDMS metric exhibits limited discriminative capacity, even with no variation in Comfortable submetric}
  \setlength{\tabcolsep}{1mm}
  \begin{tabular}{l|cc|cc}
    \hline
    \hline
    \multirow{2}{*}{Attribute} & \multicolumn{2}{c|}{Original PDMS} & \multicolumn{2}{c}{SM-PDMS} \\
    \cline{2-5}
    & std. & range & std. & range \\
    \hline
    EP       & 1.657 & 3.28 & \textbf{4.339} & \textbf{7.57} \\
    TTC      & 0.643 & 1.16 & 0.575 & 1.08 \\
    Comf.     & 0.000 & 0.00 & \textbf{0.312} & \textbf{0.59} \\
    \hline
    Scores  & 1.614 & 2.48 & \textbf{1.660} & \textbf{3.01} \\
    \hline
    \hline
  \end{tabular}
  \label{tab:sm_pdms_std_range}
\end{table}

\begin{figure*}[htbp]
  \centering
  \includegraphics[width=1.0\linewidth]{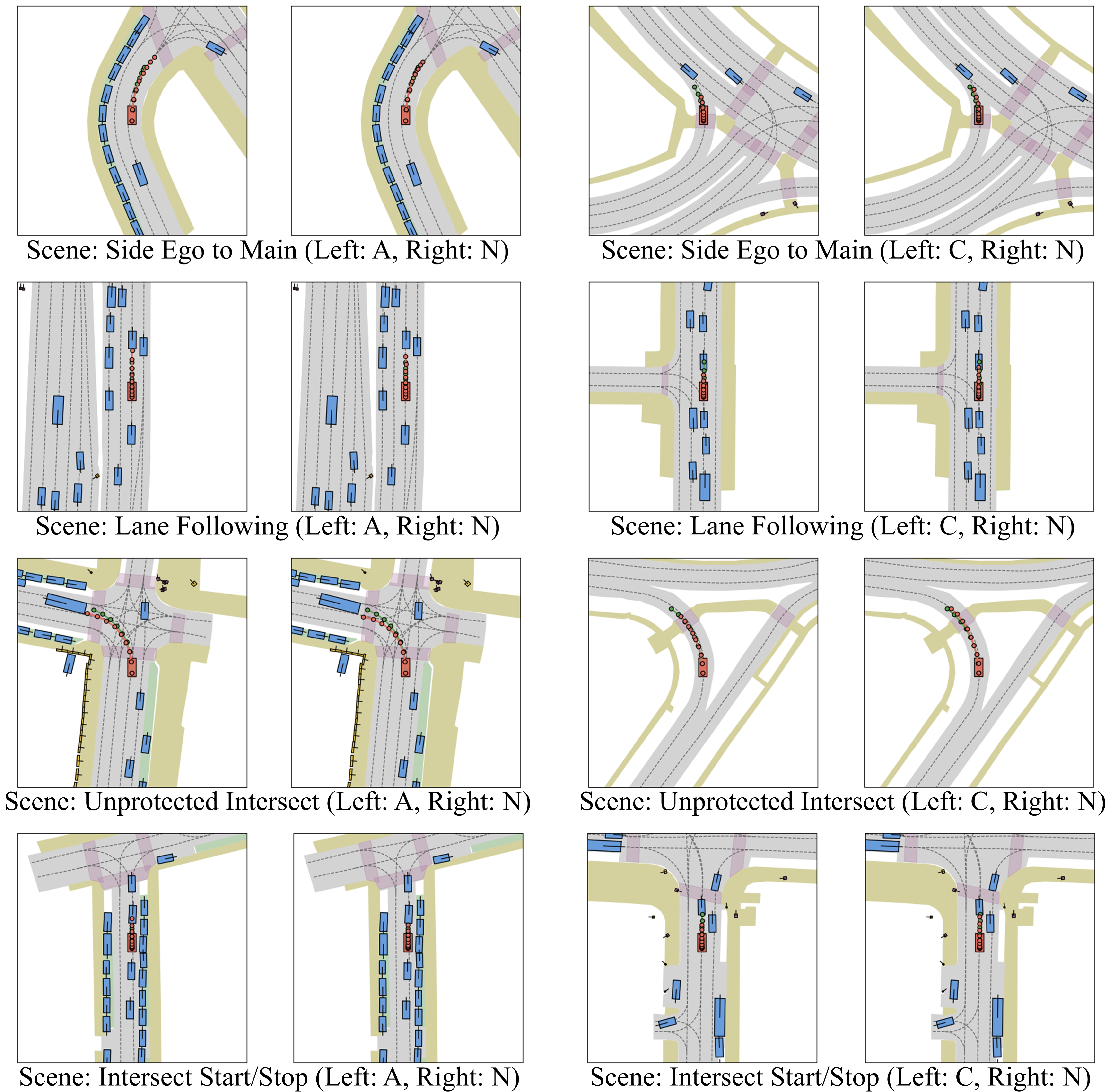}
  \caption{Qualitative illustration of DiffusionDrive-Style predictions under different style conditions across identical scenarios. Left: Aggressive (A) vs. Normal (N); Right: Conservative (C) vs. Normal (N). Red lines indicate the model's predicted trajectory under the given style condition; green lines denote the ground-truth human trajectory. Clear behavioral differences emerge with style variation, reflecting the model’s ability to adapt its outputs to driving preferences.}
  \label{fig:policy_cases}
\end{figure*}

\paragraph{Qualitative Case Study of Style Effects.}
To further illustrate the behavioral influence of style conditioning, Fig.~\ref{fig:policy_cases} shows how our DiffusionDrive-Style model produces different trajectory predictions under aggressive, conservative, and normal style inputs across identical scenarios. Across scenarios, clear differences emerge in the model's trajectory choices. These visualizations confirm that the same policy network, when conditioned on distinct style vectors, can produce behaviorally diverse outputs aligned with human-like style variations. This evidences the controllability and expressiveness afforded by the style-conditioning mechanism.

\section{Conclusion}
This paper introduces StyleDrive, a novel large-scale dataset and benchmark tailored for advancing personalized E2EAD. By systematically integrating map topology analysis, fine-grained semantic context from vision-language models, and a hybrid annotation pipeline combining rule\&distribution-based heuristics with subjective VLM-based reasoning, we establish a rich and interpretable foundation for driving style annotation. This dataset covers a broad spectrum of real-world traffic scenarios annotated with carefully designed style preference labels. To facilitate development and evaluation of personalized E2EAD, we further propose the StyleDrive Benchmark, a non-reactive simulation-based evaluation playground. Central to this benchmark is the Style-Modulated PDMS metric, which augments traditional safety and feasibility assessments with stylistic alignment measures calibrated to intended driver preferences. Extensive experiments across multiple model paradigms demonstrate the effect of style conditioning in enhancing behavioral alignment while preserving core driving competence.

\paragraph{Outlook.} \textit{\textbf{Coarse-to-Fine Style Labeling}}: Current style annotation adopts a three-level hierarchy, yet attempts at finer levels often result in ambiguous or overlapping. Further enhanced scene understanding/modeling and larger datasets may help resolve such ambiguities. \textit{\textbf{Model}}: Beyond benchmark's baselines, future work should explore joint modeling of scene context and driving style preferences. \textit{\textbf{Application}}: Inferring styles from user (long-term) profiles in real-world settings remains an open challenge with significant implications for commercial deployment.

% \newpage
\bibliographystyle{plainnat}
\bibliography{reference}

% \newpage
% \section*{NeurIPS Paper Checklist}
% \input{sec/checklist}

\newpage

\setcounter{section}{0}

\section*{Supplementary Material Overview}
This appendix provides additional experiments and detailed information related to the StyleDrive Dataset and Benchmark. Specifically:
\begin{itemize}[leftmargin=1em, topsep=0pt, itemsep=0.3em]
    \item Sect.~\ref{sec: metric} provides further details about the design of the style-aware evaluation metric.
    \item Sect.~\ref{sec: data_stat} visualizes more demos for style annotation results and illustrates additional dataset statistics.
    \item Sect.~\ref{sec: map_modify} describes modifications to map topology aimed at supporting more diverse static driving scenarios.
    \item Sect.~\ref{sec: rule_dist_ana} explains the preference annotation process using rule\&distribution-based heuristics, along with related analysis.
    \item Sect.~\ref{sec: vlm} discusses the details of benchmark methods training and VLM fine-tuning for driving scene understanding.
\end{itemize}

\section{Details about Style-aware Metric Design}\label{sec: metric}
This section provides a detailed explanation of how the Style-Modulated PDM Score (SM-PDMS) adjusts key sub-metrics: Comfort, Ego Progress, and Time-to-Collision on annotated driving styles: aggressive, normal, and conservative.

The adjustments ensure that the evaluation respects not only safety and feasibility, but also style-specific driving preferences. Below are the precise configurations used for each sub-score:

\subsection{Comfort (Comf.).}
\paragraph{Limitation of the Original Comfort Metric in PDMS}
The original PDMS framework~\cite{dauner2024navsim} evaluates comfort by penalizing trajectories exhibiting excessive acceleration and jerk, relying on fixed thresholds derived from empirical studies or simulator dynamics. While this approach effectively discourages physically unstable motions, it applies a uniform standard to all driving styles, disregarding the fact that perceived comfort can vary substantially across driver types. For instance, what a conservative driver may consider abrupt, an aggressive driver might deem acceptable. By enforcing a static threshold regardless of context or preference, the metric overlooks stylistic nuance and risks penalizing behavior that is intentionally assertive yet still within acceptable human tolerances.

\paragraph{Style-Aware Comfort Thresholding in SM-PDMS}
To better account for individual driving style preferences, we introduce style-aware thresholding for the comfort metric in SM-PDMS. Specifically, we modulate the baseline acceleration and jerk thresholds according to the designated driving style category:
\begin{itemize}[leftmargin=1em, topsep=0pt, itemsep=0.3em]
\item \textbf{Conservative:} Lower tolerance for dynamic changes. Baseline thresholds are tightened by 20\%, resulting in stricter penalization of sudden acceleration and jerk.
\item \textbf{Aggressive:} Higher tolerance for dynamic maneuvers. Thresholds are relaxed by 20\%, allowing more assertive behavior without penalty.
\item \textbf{Normal:} Retains the original default thresholds, providing a neutral baseline.
\end{itemize}

These style-adjusted thresholds govern whether a given action is penalized as uncomfortable or accepted as smooth, aligning comfort assessment with subjective human preferences. This reformulation enables the comfort metric to reflect not only physical feasibility but also stylistic intent, ensuring agents are evaluated fairly under diverse driving styles.

\subsection{Ego Progress (EP)}
\paragraph{Limitation of the Original EP Metric in PDMS}
In the original PDMS framework~\cite{dauner2024navsim}, the Ego Progress (EP) metric measures how far the ego vehicle advances along its predicted trajectory. This is done by comparing the evaluated agent’s cumulative progress to that of a fixed rule-based baseline agent (PDM)~\cite{dauner2023parting}. If the agent travels less distance than the PDM, the score is computed as the ratio of its progress to that of the PDM; otherwise, it is assigned a perfect score of 1. While this approach captures task completion in a basic sense, it introduces several limitations. First, it reduces progress evaluation to a binary assessment of whether an agent outperforms a single rule-based baseline. Second, it provides no notion of “how much” deviation is acceptable or stylistically consistent—agents that accelerate too aggressively or hesitate excessively are not penalized, as long as they surpass the PDM in distance. As a result, the original EP metric lacks sensitivity to human-like driving intent and cannot distinguish between stylistically misaligned trajectories.

\paragraph{Style-Aware Reformulation EP in SM-PDMS}
To overcome these limitations, we propose a human-referenced and style-aware reformulation of the EP metric in SM-PDMS. Instead of comparing the evaluated agent’s progress against a rule-based system, we directly compare it to a human demonstrator’s trajectory, thereby capturing both magnitude and style of motion. Let $EP_{agent}$ denote the cumulative distance traveled by the agent over a 4-second horizon, and $EP_{target}$ the distance covered by the human demonstration over the same interval. The final EP score is defined as:

\begin{equation}
S_{EP} = \max\left(1 - \alpha \times \left(\frac{EP_{agent} - EP_{target}}{Ref}\right)^2, 0\right)
\end{equation}

\noindent where $\alpha$ is a sensitivity scaling factor (empirically set to 1.2), and $Ref$ is a tolerance parameter that modulates how much deviation is considered acceptable. $Ref$ is assigned based on the human trajectory length: 3 if $EP_{target} < 10$, 5 if $10 \leq EP_{target} < 24$, 6 if $24 \leq EP_{target} < 40$, and 7 otherwise. This formulation penalizes both over- and under-progress, encouraging agents to match not only the goal-reaching behavior but also the progressive characteristics of human driving, such as decisiveness, hesitation, or assertiveness. In doing so, it aligns the metric more closely with the stylistic fidelity goals of SM-PDMS.

\subsection{Time-to-Collision (TTC)}

\paragraph{Limitation of the Original TTC Metric in PDMS}
In its original implementation, the Time-to-Collision (TTC) metric adopts a binary risk assignment scheme. Specifically, a TTC score of 0 is assigned to a trajectory once any potential geometric intersection is detected between the ego vehicle and another traffic participant at any future time step, provided that certain semantic conditions are met—such as the interacting agent being located ahead of the ego vehicle, or the ego being in a complex or non-drivable area (e.g., an intersection or multiple lanes). Otherwise, the TTC score remains at its default value of 1.

This approach effectively reduces the TTC computation to a binary classification of “collision risk” versus “no risk,” without incorporating the actual temporal distance to the potential collision. That is, whether a collision is predicted to occur in 0.1 seconds or 1.4 seconds, it is treated identically under the original design. As a result, the method neglects the core intuition of TTC as a temporal measure of urgency, thereby limiting its ability to differentiate trajectories by the imminence of risk.

\paragraph{Style-Aware Temporal-Thresholding TTC in SM-PDMS}
To address mentioned limitation, we propose a revised TTC scoring mechanism that explicitly incorporates the predicted time until collision (collision time) and compares it against a driving-style-aware threshold. For each trajectory proposal, we record the earliest future time step at which a potential collision is detected. This time-to-collision value is then compared against a predefined threshold determined by the assumed driving style: for instance, 0.8 seconds for aggressive drivers, 1.0 second for neutral drivers, and 1.2 seconds for conservative drivers.

A binary score is then assigned based on this comparison:
\begin{itemize}
    \item If the collision time is greater than or equal to the threshold, the trajectory is considered acceptable and assigned a score of 1;
    \item Otherwise, it is deemed unsafe and assigned a score of 0.
\end{itemize}

This approach preserves the simplicity of binary scoring while introducing temporal sensitivity and personalized risk tolerance, thereby aligning the TTC metric more closely with style preference and human driving behaviors.

\subsection{Othe submetrics, Aggregation, and Notes}

Other safety-related submetrics include NC and DAC defined in Navsim~\cite{dauner2024navsim} are retained, and we also retain Navsim's aggregation method for all submetrics.

The PDMS referenced throughout the paper and its appendix corresponds to the Navsim~\cite{dauner2024navsim} version 1 implementation.

\section{More Demo Visualization and Illustration of Dataset Statistics}\label{sec: data_stat}

\subsection{Demo Visualization for Style Annotation Results}

\begin{figure*}[htbp]
    \centering
    \begin{subfigure}[b]{1.0\textwidth}
        \centering
        \includegraphics[width=0.75\textwidth]{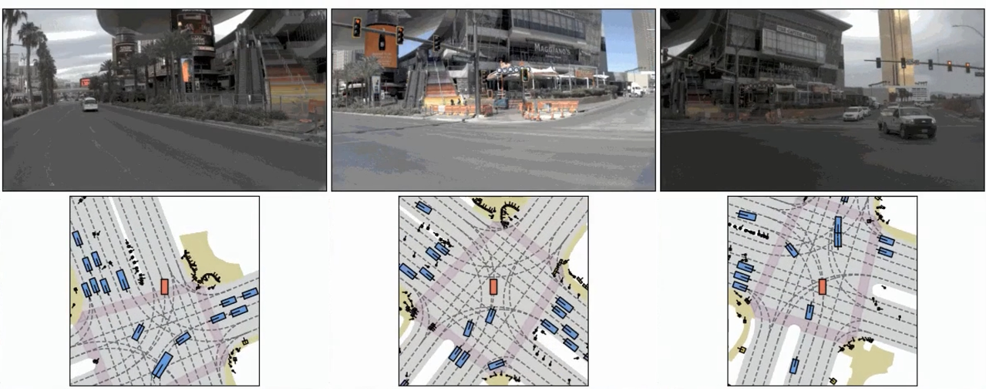}
        \caption{Given similar initial conditions for a \textbf{left-turn maneuver}, after 7 seconds, the aggressive driver (left) exhibits higher acceleration and covers the greatest progress, the normal driver (center) progresses moderately toward the intersect center, and the conservative one (right) initiates movement slowly, leading to mini displacement.}
        \label{fig:left_turn}
    \end{subfigure}
    \hfill
    \begin{subfigure}[b]{1.0\textwidth}
        \centering
        \includegraphics[width=0.75\textwidth]{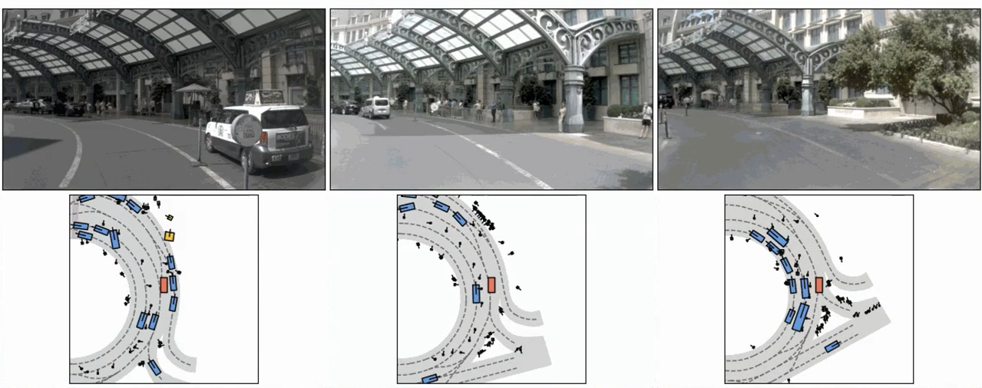}
        \caption{Given similar initial conditions for an \textbf{interior-road scenario}, after 7 seconds, the aggressive driver (left) exhibits higher speed and covers the greatest progress, ignoring potential sudden pedestrian emergence or vehicle merging in, the normal driver (center) progresses moderately, and the conservative one (right) moves slowly and cautiously in this complex scenario.}
        \label{fig:interior_road}
    \end{subfigure}
    \hfill
    \begin{subfigure}[b]{1.0\textwidth}
        \centering
        \includegraphics[width=0.75\textwidth]{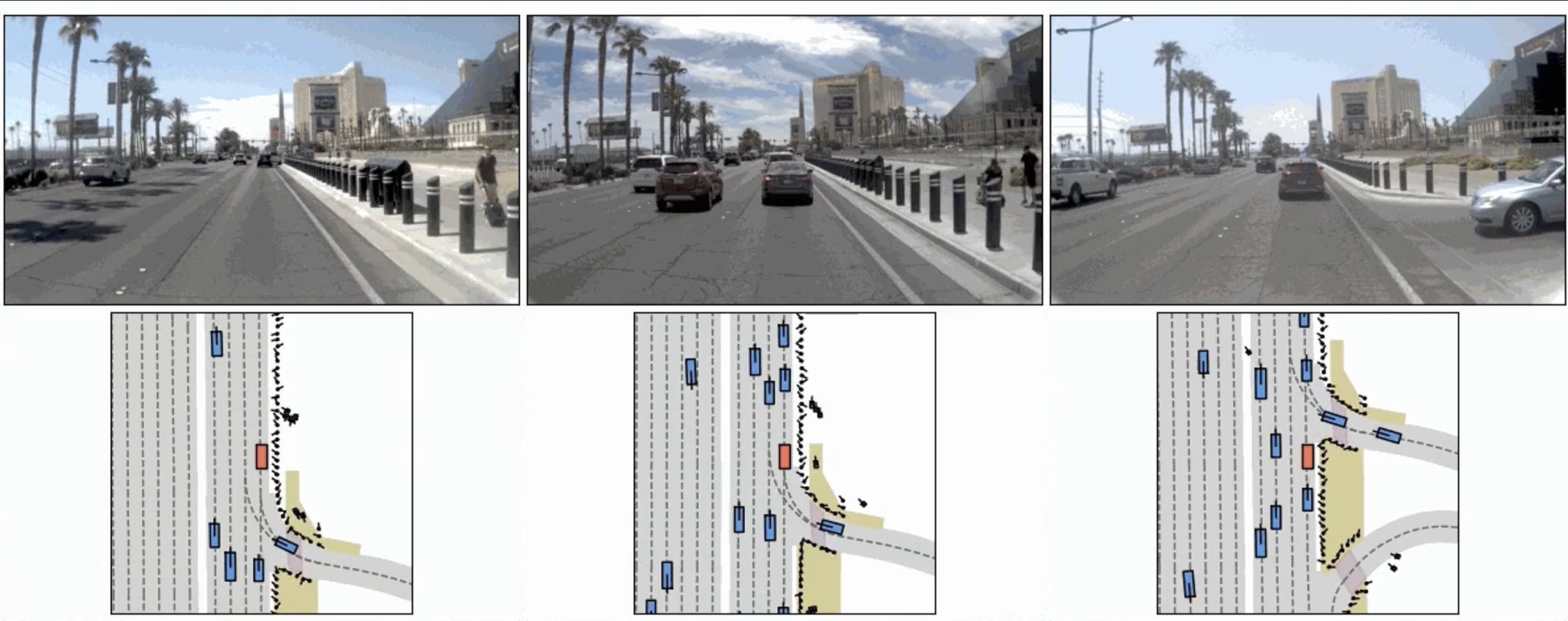}
        \caption{Given similar initial conditions for a \textbf{side-to-main scenario} (with ego on the main road), after 7 seconds, the aggressive driver (left) ignores the merging vehicle and maintains high speed to pass the merge point; the normal driver (center) decelerates moderately but still passes before the merging vehicle; whereas the conservative driver (right) maintains a large gap from the lead vehicle and even brakes to yield space for the merging vehicle.}
        \label{fig:right_turn}
    \end{subfigure}
    \caption{ Visualization of distinct driving behaviors from similar initial positions in three representative scenarios. The trajectories reflect clearly distinguishable aggressive (left), normal (center), and conservative (right) driving tendencies, demonstrating the validity and diversity of the annotated style labels.}
    \label{fig:video_demos}
\end{figure*}

To illustrate the validity and diversity of the style annotations in our dataset, we visualize distinct driving behaviors from similar initial positions across three representative scenarios. As shown in Fig.~\ref{fig:video_demos}, the vehicles exhibit clearly distinguishable aggressive, normal, and conservative driving tendencies. 

It's worth noting that these meaningful style distinctions are enabled by our comprehensive modeling and annotation framework, as well as the inherent richness of the original OpenScene dataset, which was collected from vehicles operated by different drivers across diverse and complex urban environments in both the USA and Singapore.

\subsection{Illustration of Dataset Statistics}

\begin{figure*}[htbp]
    \centering
    \begin{subfigure}[b]{0.32\textwidth}
        \includegraphics[width=\textwidth]{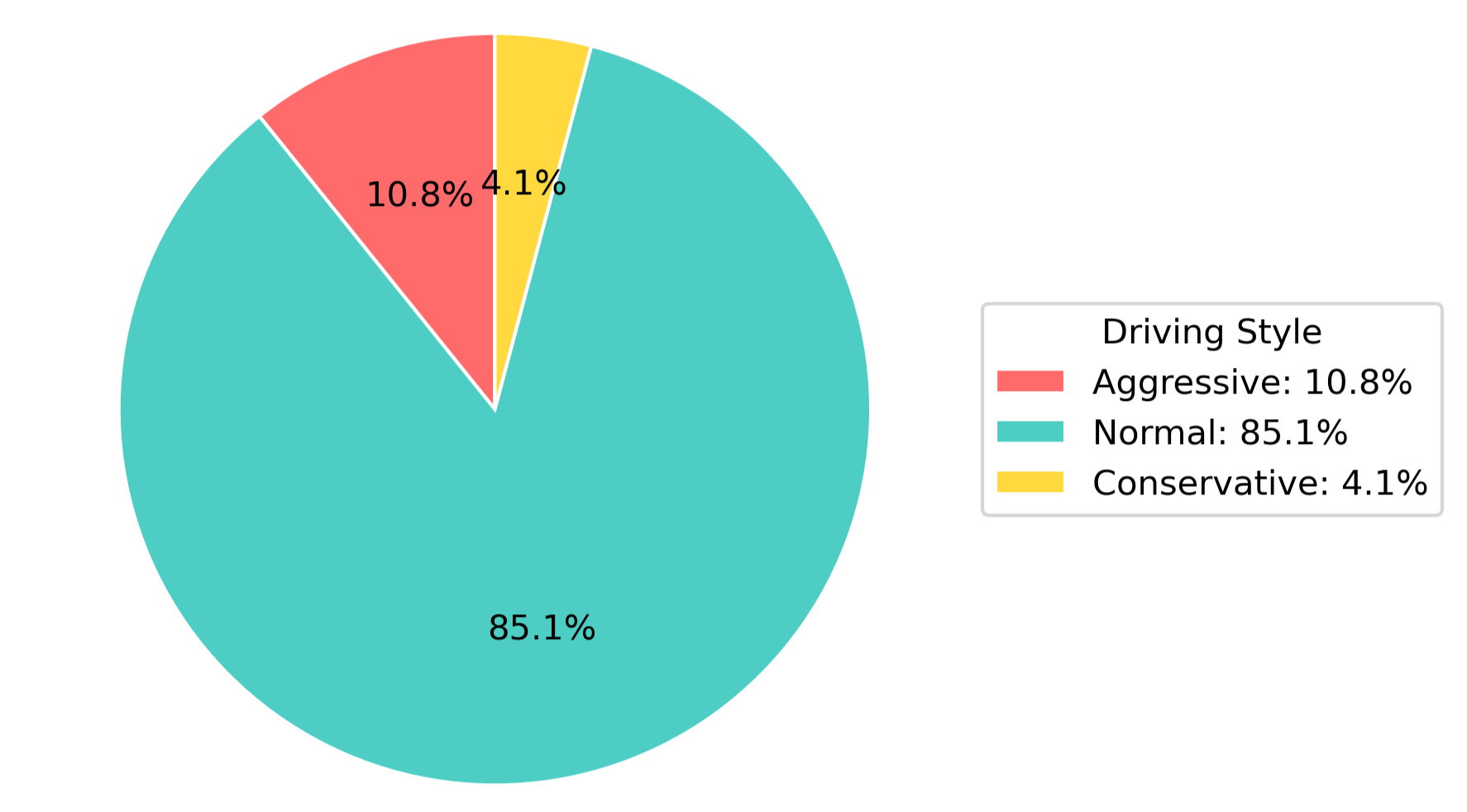}
        \caption{Lane Following}
        \label{fig:pie_lane_following}
    \end{subfigure}
    \hfill
    \begin{subfigure}[b]{0.32\textwidth}
        \includegraphics[width=\textwidth]{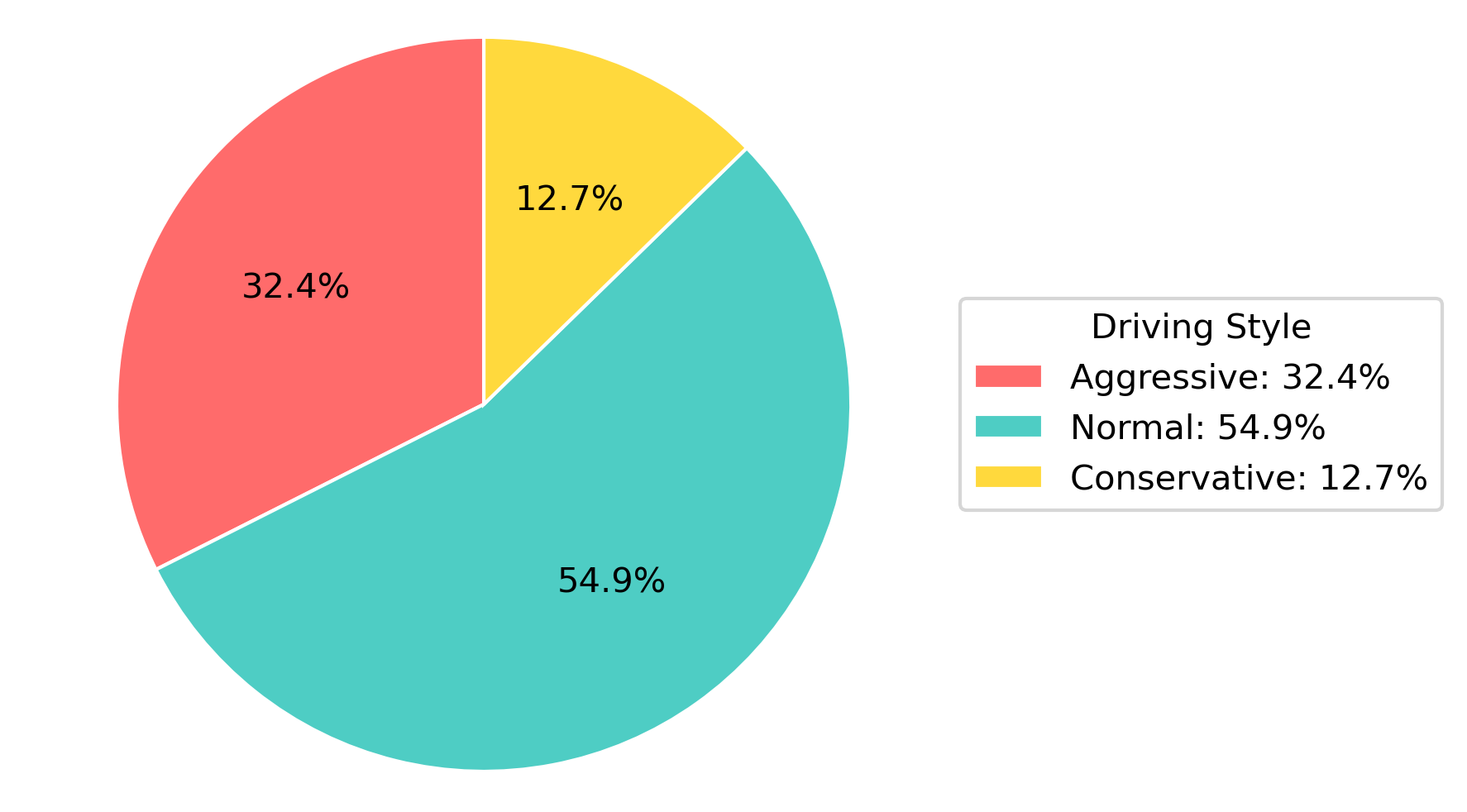}
        \caption{Lane Change}
        \label{fig:pie_lane_change}
    \end{subfigure}
    \hfill
    \begin{subfigure}[b]{0.32\textwidth}
        \includegraphics[width=\textwidth]{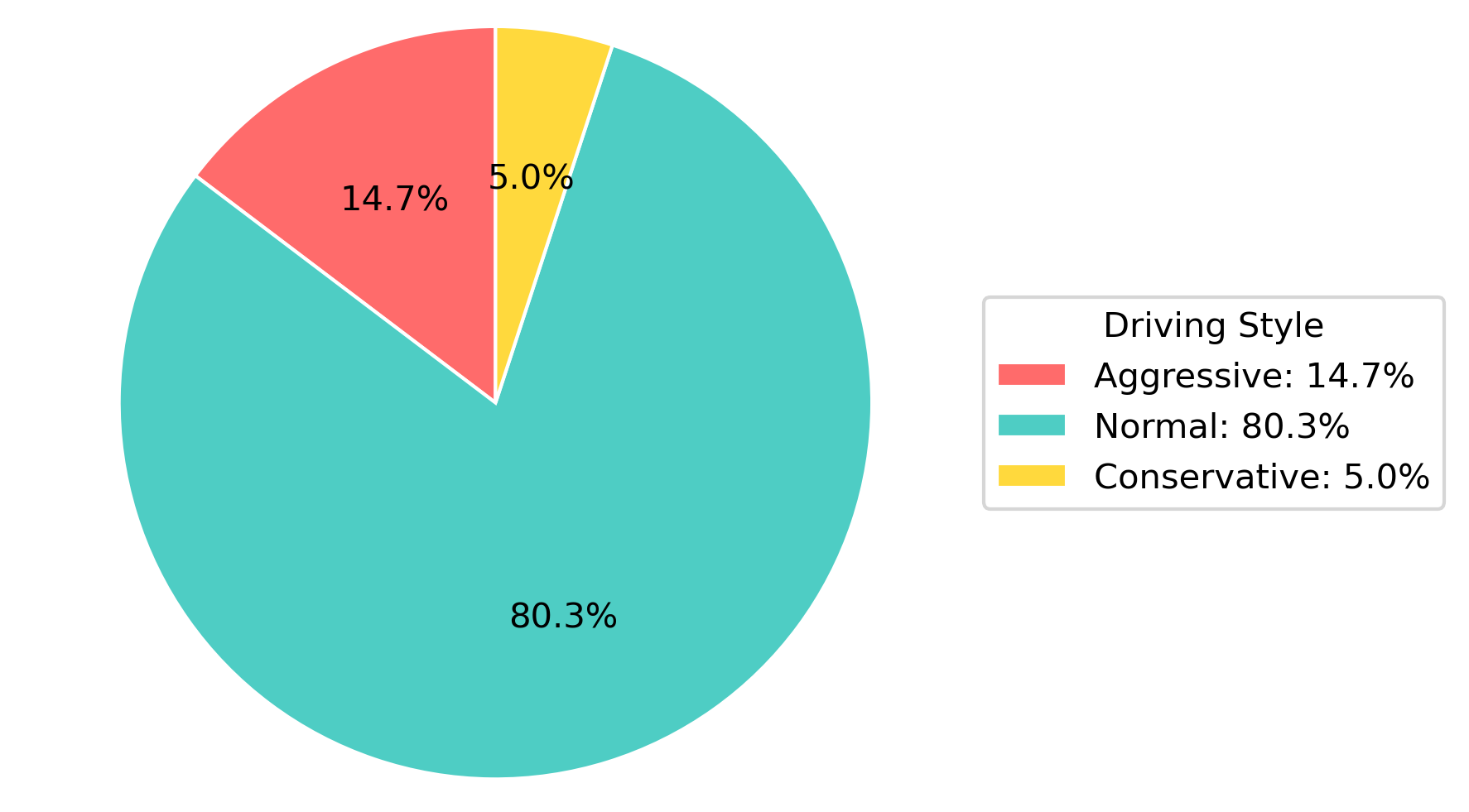}
        \caption{Crosswalk}
        \label{fig:pie_crosswalk}
    \end{subfigure}
    
    \vspace{0.8em}

    \begin{subfigure}[b]{0.32\textwidth}
        \includegraphics[width=\textwidth]{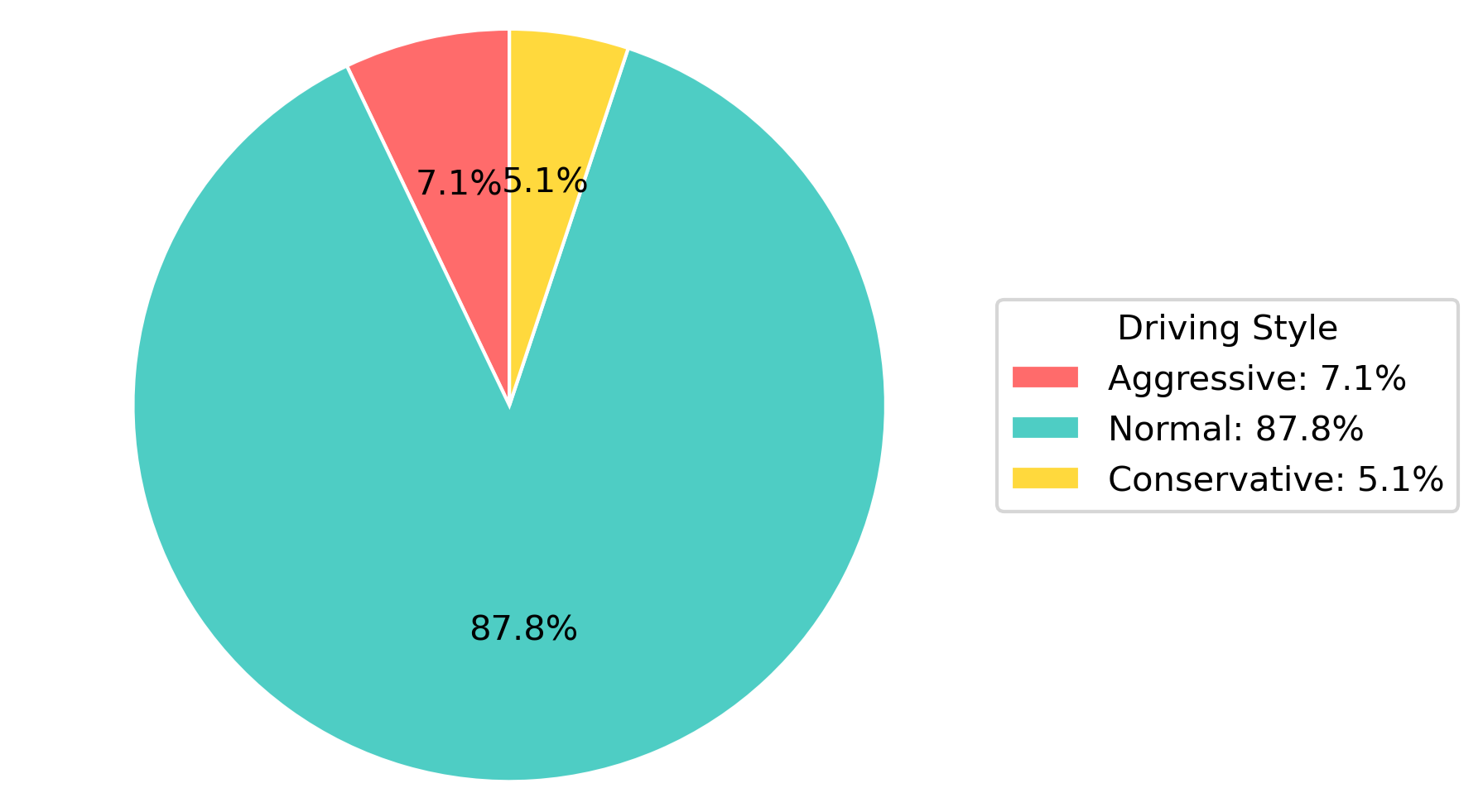}
        \caption{Side-to-Main (Ego on Main Road)}
        \label{fig:pie_side_to_main_main}
    \end{subfigure}
    \hfill
    \begin{subfigure}[b]{0.32\textwidth}
        \includegraphics[width=\textwidth]{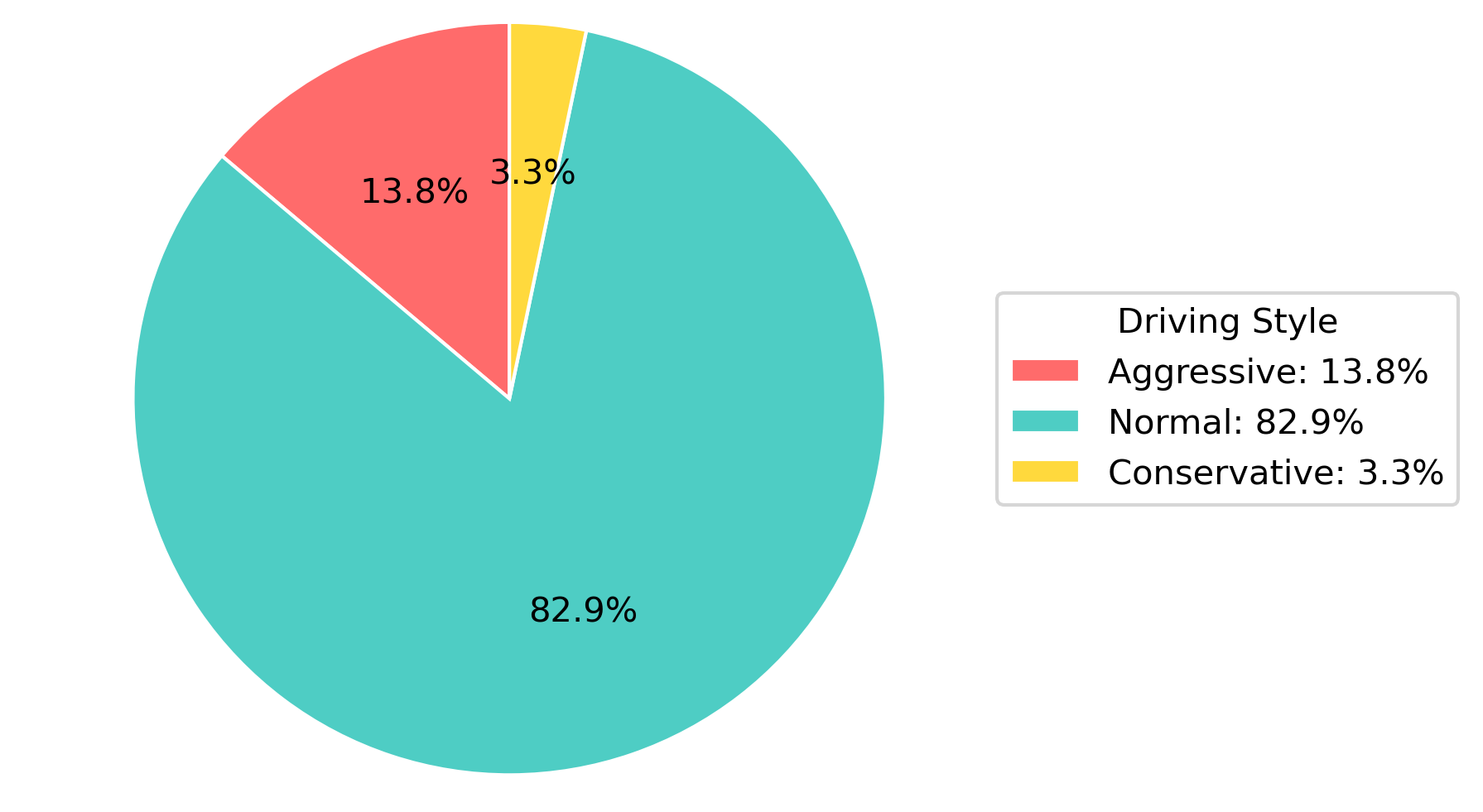}
        \caption{Side-to-Main (Ego on Side Road)}
        \label{fig:pie_side_to_main_side}
    \end{subfigure}
    \hfill
    \begin{subfigure}[b]{0.32\textwidth}
        \includegraphics[width=\textwidth]{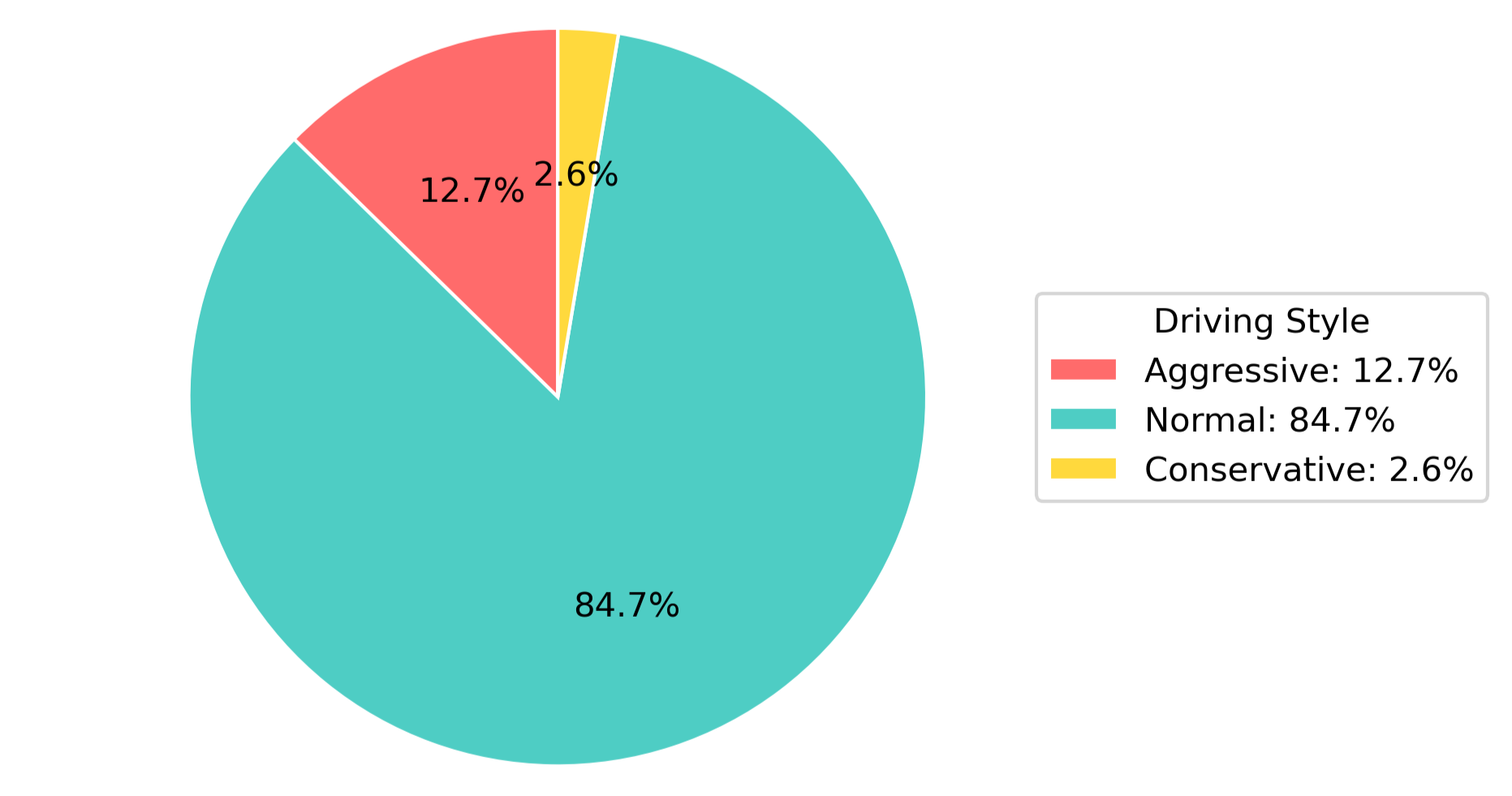}
        \caption{Protected Intersections}
        \label{fig:pie_protected_intersections}
    \end{subfigure}

    \vspace{0.8em}

    \begin{subfigure}[b]{0.32\textwidth}
        \includegraphics[width=\textwidth]{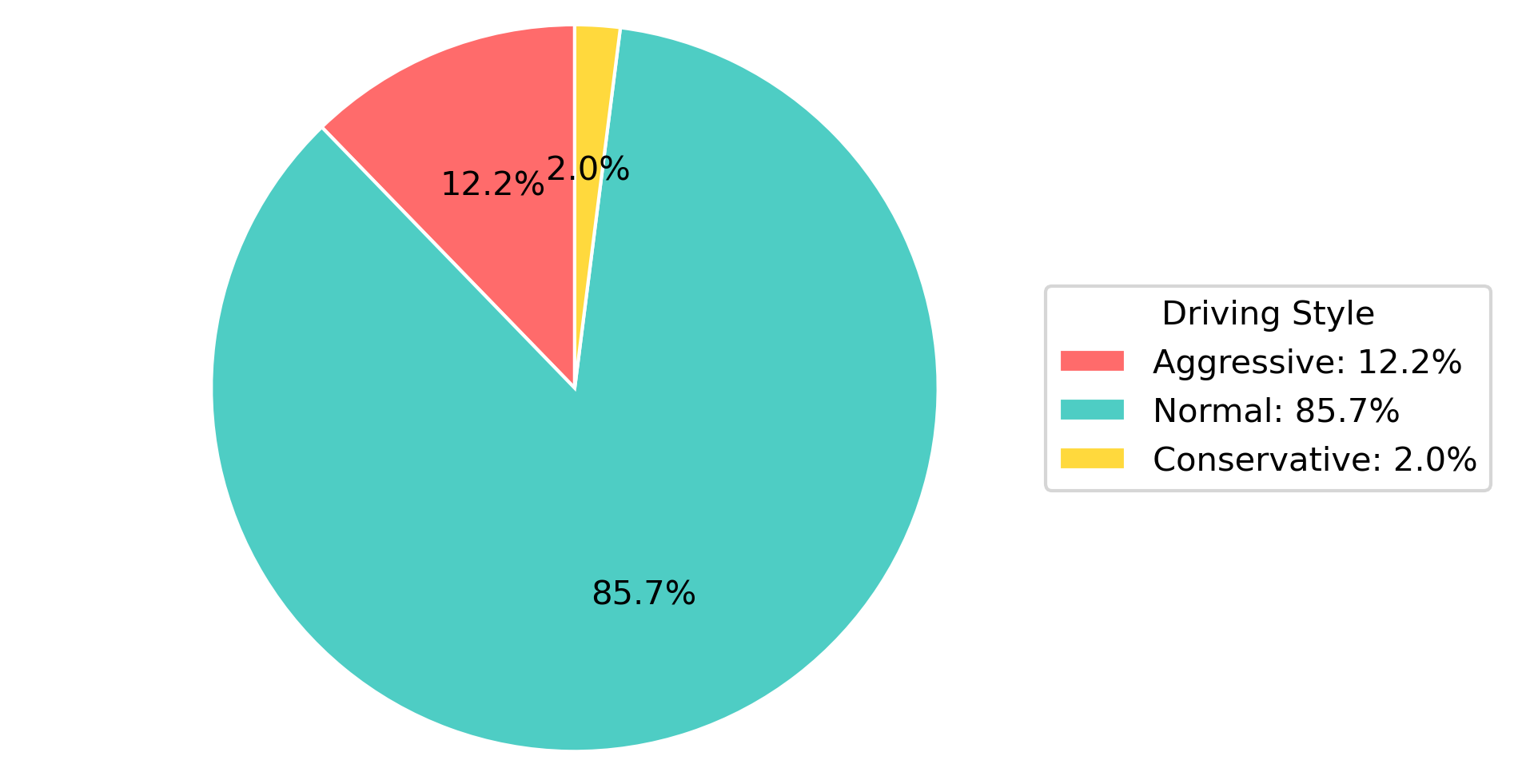}
        \caption{Unprotected Intersections}
        \label{fig:pie_unprotected_intersections}
    \end{subfigure}
    \hfill
    \begin{subfigure}[b]{0.32\textwidth}
        \includegraphics[width=\textwidth]{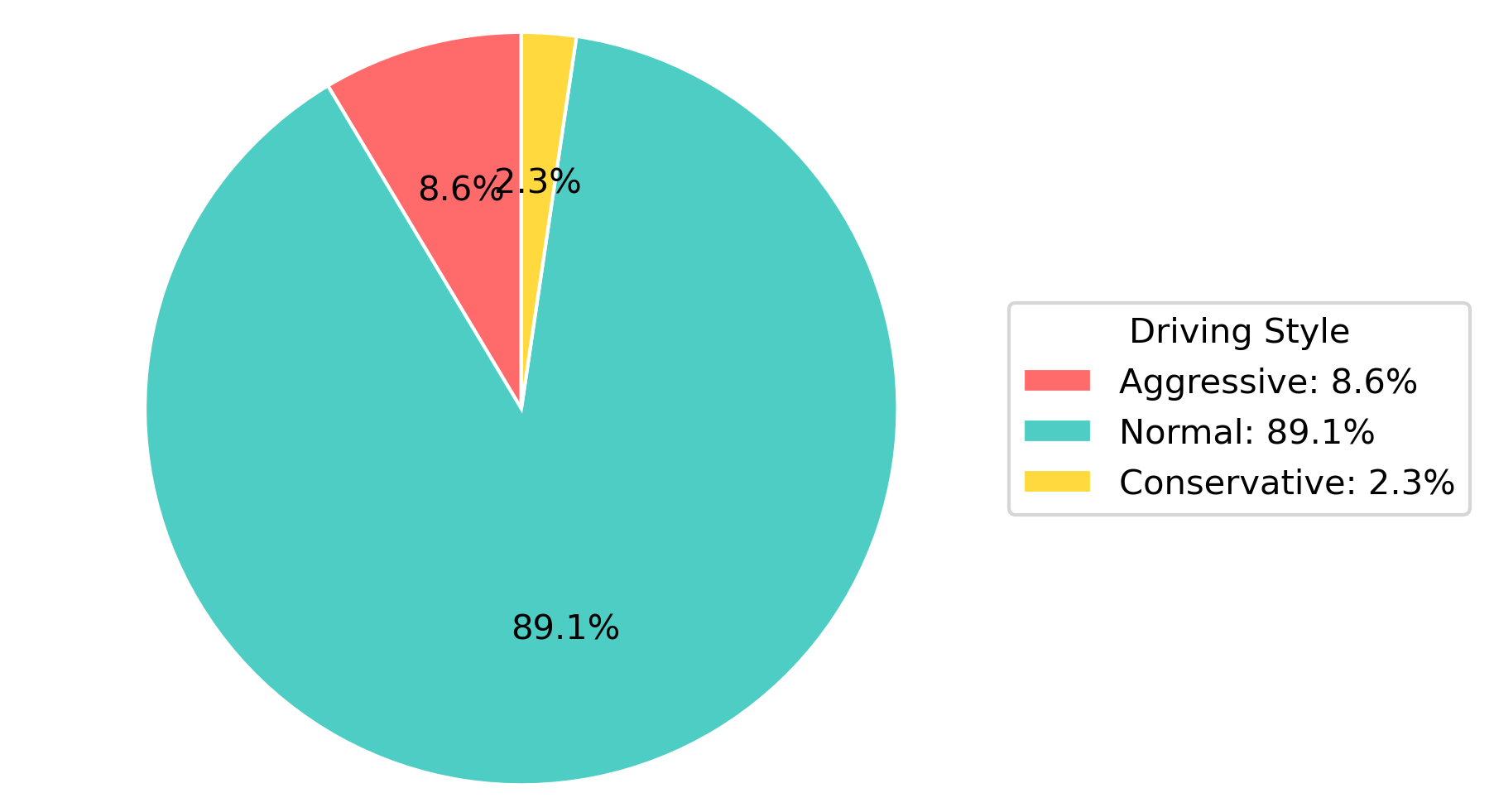}
        \caption{Roundabout Entrance}
        \label{fig:pie_roundabout_entrance}
    \end{subfigure}
    \hfill
    \begin{subfigure}[b]{0.32\textwidth}
        \includegraphics[width=\textwidth]{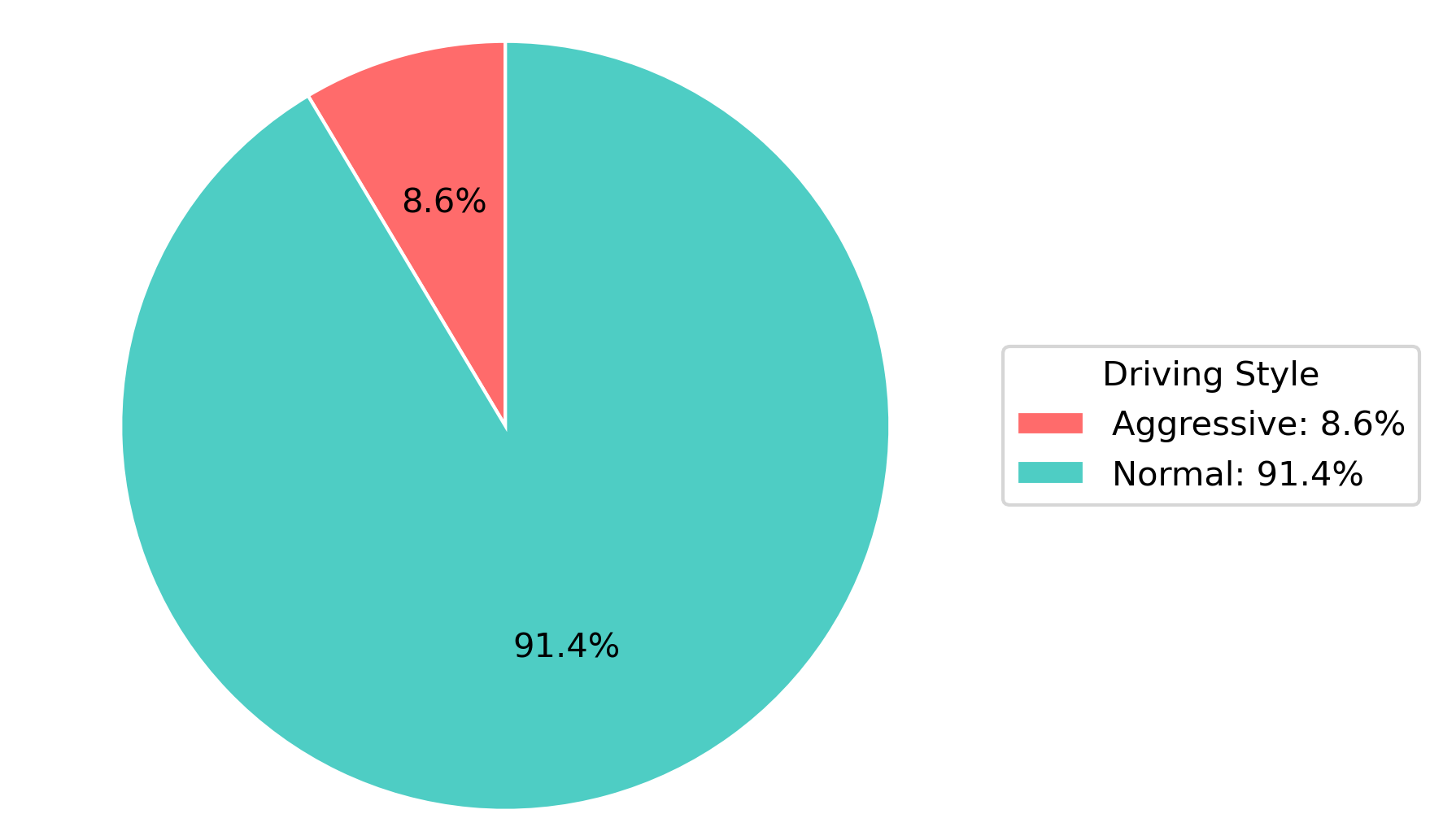}
        \caption{Roundabout Interior}
        \label{fig:pie_roundabout_interior}
    \end{subfigure}
    
    \vspace{0.8em}
    
    \begin{subfigure}[b]{0.32\textwidth}
        \includegraphics[width=\textwidth]{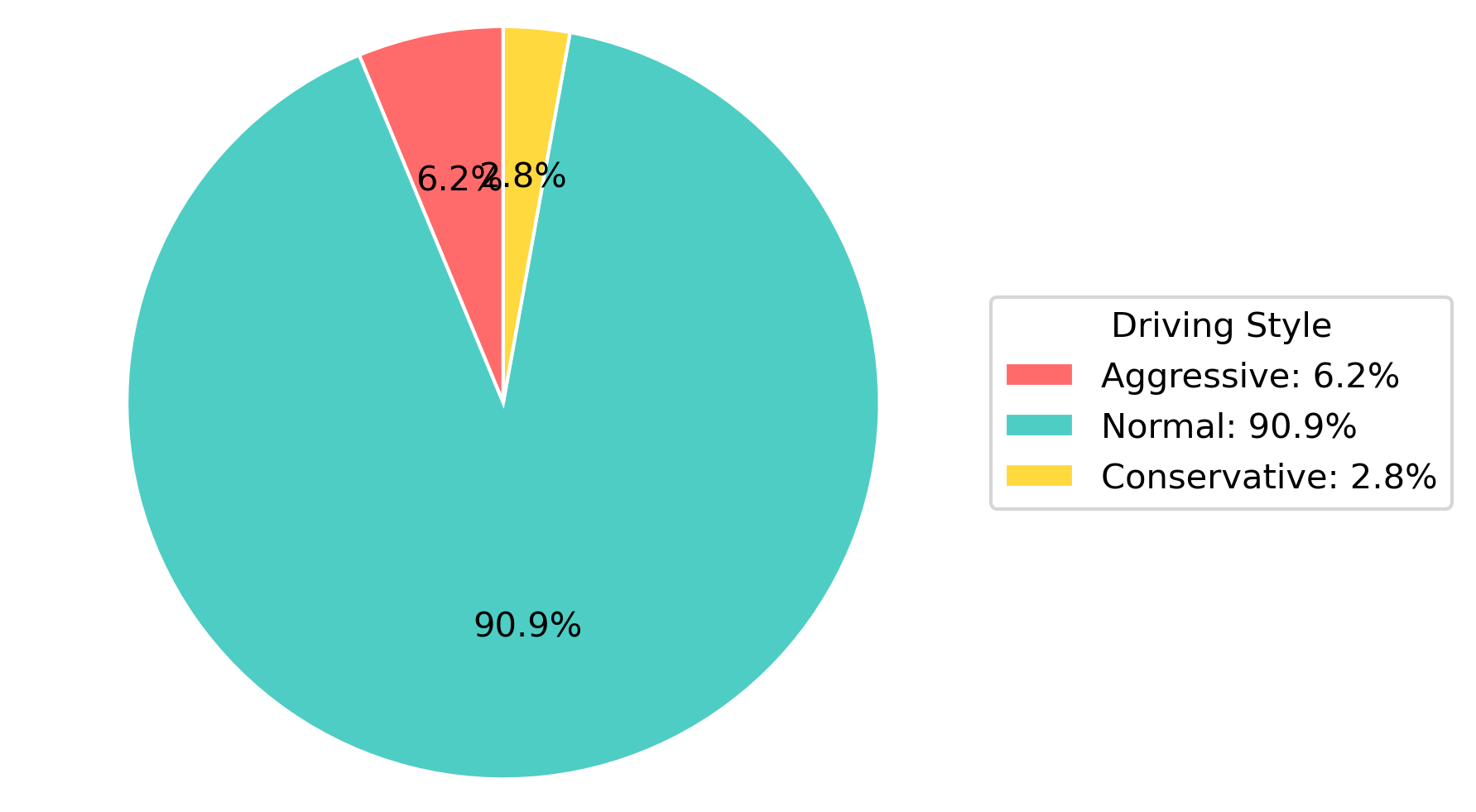}
        \caption{Countryside Road}
        \label{fig:pie_countryside}
    \end{subfigure}
    \hfill
    \begin{subfigure}[b]{0.32\textwidth}
        \includegraphics[width=\textwidth]{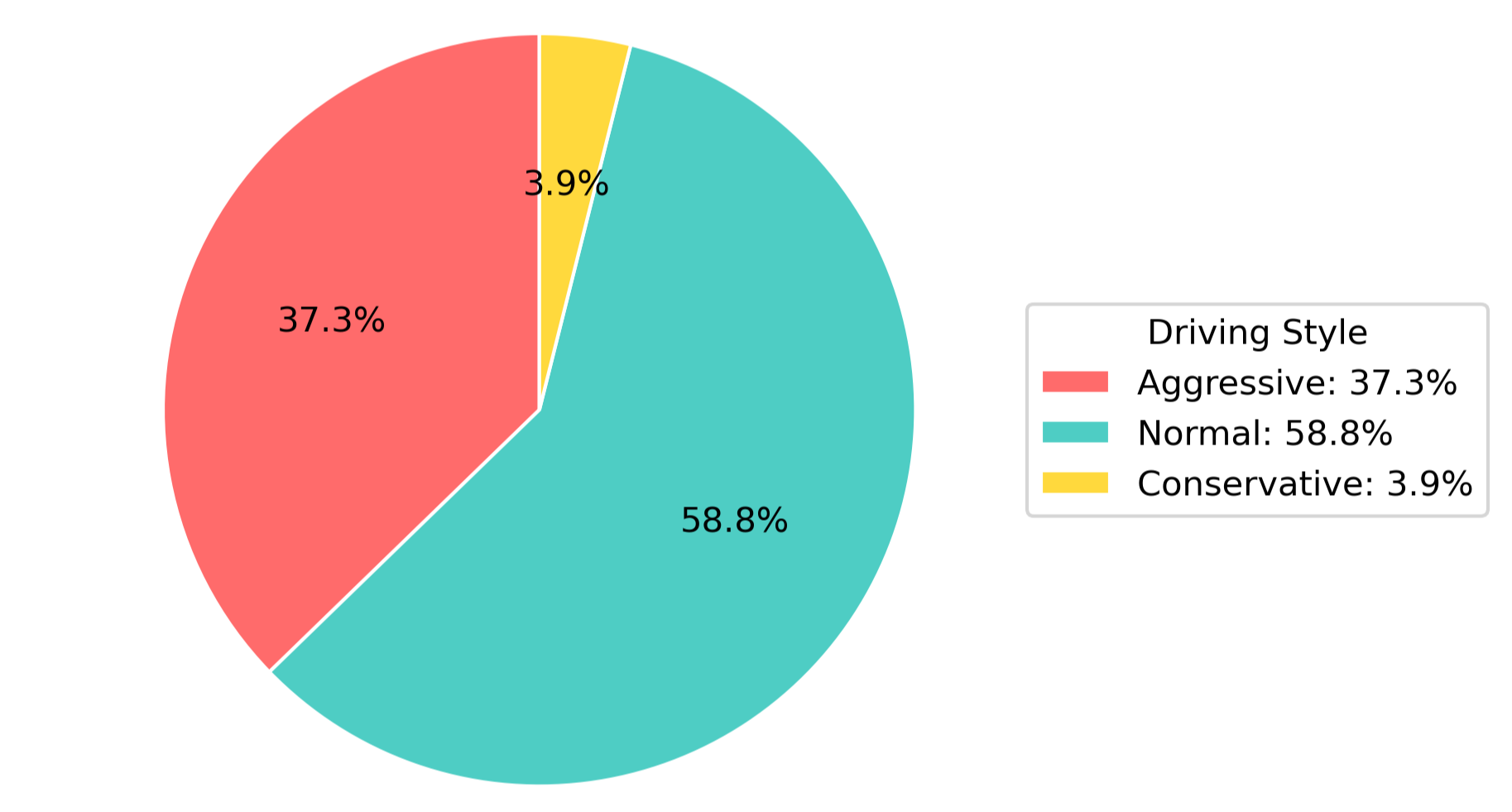}
        \caption{Special Interior}
        \label{fig:pie_special}
    \end{subfigure}
    
    \caption{Per-scenario distribution of annotated driving styles (A: Aggressive, N: Normal, C: Conservative).}
    \label{fig:all_pie_charts}
\end{figure*}

To provide a comprehensive overview of driving style annotations, we visualize the distribution of \textit{Aggressive (A)}, \textit{Normal (N)}, and \textit{Conservative (C)} labels across key scenario types. Each pie chart summarizes the proportion of styles within a specific behavioral context. Fig.~\ref{fig:pie_lane_following}-\ref{fig:pie_unprotected_intersections} present the distributions across 11 fine-grained scenario categories.

Lane-following scenes are primarily dominated by Normal behavior (85.1\%), with Aggressive and Conservative styles accounting for 10.8\% and 4.1\%, respectively. Lane changes show a relatively higher proportion of Aggressive behavior (32.4\%) and Conservative behavior (12.7\%), suggesting diverse preferences in lateral maneuvers. In crosswalk scenarios, Normal behavior still dominates (80.3\%), but with notable instances of Aggressive (14.7\%) and Conservative (5.0\%) driving.

In merging-related scenes, side-to-main (main road) shows 7.1\% Aggressive and 5.1\% Conservative behavior, while side-to-main (side road) records 13.8\% Aggressive and 3.3\% Conservative behavior, with 82.9\% remaining Normal. Protected intersections contain 12.7\% Aggressive behavior, slightly higher than the 12.2\% observed in unprotected intersections, which also exhibit 2.0\% Conservative driving.

At roundabouts, entrance segments exhibit 8.6\% Aggressive behavior and 2.3\% Conservative behavior, while interior segments are more stable, with 91.4\% Normal and 8.6\% Aggressive. Special interior roads show higher variability, including 37.3\% Aggressive and 3.9\% Conservative behavior. Countryside roads remain generally stable, with 90.9\% Normal and only 6.2\% Aggressive driving observed. In carpark areas, behavior is entirely Normal (100.0\%).

\section{Map Topology Modifications towards More Diverse Scenarios}\label{sec: map_modify}
To enhance the accuracy of static scene classification, we manually refine the NuPlan HD map using QGIS. These adjustments target known limitations in the original topology and provide a more semantically meaningful foundation for behavior modeling.

\paragraph{Refined Intersection Semantics.}
The original intersection definitions are overly coarse. We introduce fine-grained polygonal annotations for:
\begin{itemize}[leftmargin=1em, topsep=0pt, itemsep=0.3em]
    \item Merging Zones: areas where minor roads (e.g., side streets or on-ramps) merge into a major road. These regions often pose visibility challenges and merging-in risks.
    \item Main Road Merge Receptive Zones: areas along major roads where merging from adjacent lanes or secondary roads is expected. These zones carry merging-in conflicts and courtesy yield challenges.
\end{itemize}

\paragraph{Functional Area Roads.}
We add internal road polygons for areas such as casinos, hotels, hospitals, and airports. These environments feature irregular geometry, frequent stops, and dense interactions with non-standard road agents. They are often associated with low-speed navigation risks, occlusions, and high pedestrian density, making them crucial for evaluating slow-driving and cautionary styles.

\paragraph{Roundabout-Specific Regions.}
We annotate both roundabout entries and interior roundabouts. These zones are associated with misjudgment of right-of-way, incorrect merging-in, and delayed exits, posing non-trivial planning challenges.

\paragraph{Rural Roads.}
Given the distinct structure of Pittsburgh’s rural roads - e.g., narrow lanes, limited markings, and relatively sharp curves - we isolate them to better capture related driving behaviors.

\paragraph{Intersection Normalization.}
We standardize NuPlan’s intersection types by aligning definitions and separating control logic (e.g., stop-sign vs. traffic-light), enabling consistent downstream scene labeling.

\section{More Details about Preference Annotation via Rule\&Distribution-based Heuristics}\label{sec: rule_dist_ana}

To provide a transparent and robust methodology for driving style annotation, we detail the heuristic rules used to classify driving behavior across a diverse range of traffic scenarios. These rules are developed based on scenario subcategories refined through systematic scenario modeling, ensuring contextual relevance. Within each subcategory, population-level behavior distributions are analyzed to statistically derive rule thresholds. The rules themselves are grounded in interpretable, physically meaningful features that capture motion dynamics and contextual semantics, as discussed in the main paper. Final threshold calibration is further refined through expert validation, enabling scenario-specific, objective, and explainable preference labeling.

Table~\ref{tab:style-rule-summary} summarizes the dynamic features, contextual variables, and scenario types that form the basis of our rule-based driving style classification. These criteria guide the labeling across different driving situations, enabling consistency and generalizability in behavioral analysis.

\begin{table*}[ht]
\centering
\caption{Rule-based Heuristic Criteria for Driving Style Classification Across Scenarios}
\label{tab:style-rule-summary}
\begin{tabular}{p{2.5cm} p{4cm} p{6.5cm}}
\toprule
\textbf{Traffic Scenario Type} & \textbf{Contextual Variants} & \textbf{Main Dynamic Features} \\
\midrule
Lane Following & Lead distance (Close/Far/None), Road shape (Straight/Curve) & Ego speed and acceleration trend, front vehicle proximity \newline unsafe/safe frame ratio \\
\midrule
Protected Intersections & Turn direction (Left/Right/Straight), pedestrian/lead presence & Ego speed, acceleration, front gap safety ratio \newline turning intent and crossing semantics \\
\midrule
Unprotected Intersections & Similar to protected, but without signal control & Speed and yaw change pattern, gap safety evaluation \newline presence of pedestrian or lead car \\
\midrule
Lane Change & Rear vehicle presence & Lateral dynamics ($v_y$, $a_y$, yaw diff, rear threat) \\
\midrule
Special Interior Road & Lead/pedestrian presence, merging complexity & Unsafe duration, deceleration behavior \newline merge risk indicator \\
\midrule
Side to Main Ego & With/without merging risk & Ego speed trend, lateral velocity, merging maneuver assertiveness \\
\midrule
Side Ego to Main & Main road vehicle/pedestrian presence & Ego speed and acceleration under incoming vehicle threat \\
\midrule
Crosswalks & Pedestrian presence & Ego speed and acceleration trend near crosswalk \newline behavioral intent (e.g., yield or accelerate) \\
\midrule
Roundabout Entrance & With/without yield/merge risk & Ego speed and entry maneuver stability, \newline yield behavior to surrounding traffic \\
\midrule
Roundabout Interior & - & Circular stability, lateral acceleration and smoothness of ego motion \\
\midrule
Countryside Road & Road shape (Straight/Curve) & Ego speed, curvature following \newline longitudinal and lateral acceleration smoothness \\
\bottomrule
\end{tabular}
\end{table*}

We elaborate on the rule design for each scenario type in the following subsections.

\subsection{Heuristics in Lane-Following Scenarios}

In lane-following scenarios, driving style is categorized as \textit{Aggressive (A)}, \textit{Normal (N)}, or \textit{Conservative (C)} based on motion dynamics, road geometry, and interactions with leading vehicles. Our classification framework is rule\&distribution-based, utilizing both regression-derived velocity trends and context-aware features to determine driving style.

\paragraph{Velocity Trend Estimation.}
The ego vehicle's speed profile $v(t)$ is analyzed using linear and quadratic regression fits to derive trends:
\begin{itemize}[leftmargin=1em, topsep=0pt, itemsep=0.3em]
    \item \textit{Accelerating} or \textit{decelerating}: captured via linear fit with significant slope;
    \item \textit{Accelerate-then-decelerate} or \textit{decelerate-then-accelerate}: captured by non-monotonic quadratic fit;
    \item \textit{Quasi-constant}: determined when $\mathrm{std}(v(t))$ is low and no trend is dominant.
\end{itemize}

\paragraph{Contextual Modulation.}
Trend interpretation is modulated by road shape---\textit{Straight} or \textit{Curved}---and presence of a lead vehicle. For samples with a lead vehicle, we define an unsafe-following ratio to characterize risky proximity, as well as a safe-following ratio to detect cautious behavior. These thresholds are adaptively determined from empirical data statistics.

\paragraph{Rule\&Distribution-based Classification.}
Final driving style is determined based on velocity trend, average speed, peak acceleration, acceleration variability, and safety proximity ratios:
\begin{itemize}[leftmargin=1em, topsep=0pt, itemsep=0.3em]
    \item \textbf{Aggressive (A):} Assigned when unsafe-following ratios are high, or motion signals such as velocity, acceleration, or variability significantly exceed normal ranges.
    \item \textbf{Conservative (C):} Assigned when the ego vehicle maintains low speed, low acceleration variability, or consistently large gaps from lead vehicles.
    \item \textbf{Normal (N):} Assigned to samples not satisfying either extreme condition.
\end{itemize}

The rule adapts across 6 sub-conditions: [Lead: Close / Far / None] $\times$ [Road Shape: Straight / Curved]. For No Lead cases, classification relies purely on statistical motion descriptors without distance-based modulation.

\subsection{Heuristics in Special Interior Road Scenarios}

In \textit{special interior road} scenarios—characterized by narrow lanes, merging zones, or pedestrian-prone areas—driving style is determined through a combination of ego-vehicle motion trends and local traffic semantics. The rule-based framework incorporates velocity and acceleration patterns modulated by contextual indicators, including merging interactions, lead vehicles, and pedestrian presence.

\paragraph{Velocity Trend Characterization.}
The ego vehicle's longitudinal motion is analyzed by fitting the velocity sequence $v(t)$ using both linear and quadratic regression models. Based on model residuals and slope coefficients, the trend is categorized as \textit{accelerating}, \textit{decelerating}, \textit{quasi-constant}, or \textit{non-monotonic}, following the same trend extraction procedure.

\paragraph{Contextual Modulation.}
Two semantic cues modulate rule interpretation: (i) the presence of a lead vehicle; and (ii) pedestrian activity. These indicators adaptively influence thresholds derived from the statistical distribution of the dataset.

\paragraph{Rule\&Distribution-based Classification.}
Style labels are determined by combining motion descriptors and context signals:

\begin{itemize}[leftmargin=1em, topsep=0pt, itemsep=0.3em]
    \item \textbf{Merging-risk scenarios:} \textbf{Aggressive (A)} behavior is inferred when acceleration or velocity patterns are noticeably intense; \textbf{Conservative (C)} is assigned for slow and steady merging motion.
    
    \item \textbf{With lead vehicles or pedestrians:} \textbf{Conservative (C)} is assigned if the ego vehicle demonstrates cautious behavior under interaction constraints. Conversely, close following and high motion signals suggest \textbf{Aggressive (A)} behavior.
    
    \item \textbf{Isolated contexts (no leads or pedestrians):} Labels are purely determined by motion statistics, using thresholds extracted from the overall trajectory distribution in the dataset.
\end{itemize}

\subsection{Heuristics in Intersection Scenarios}

Intersection scenarios—comprising both \textit{protected} (e.g., signalized or stop-controlled) and \textit{unprotected} (e.g., unsignalized or yield-controlled) junctions—entail complex decision-making under spatial and social constraints. Driving style classification in these settings relies on vehicle maneuver intent, interaction risk with surrounding agents, and motion dynamics.

\paragraph{Maneuver Detection.}
The ego vehicle's maneuver type is inferred from its yaw change over time. Turning maneuvers (left or right) are identified by sustained directional change, while low yaw variation indicates a straight traversal. This maneuver classification influences the interpretation of motion behavior and style thresholds.

\paragraph{Interaction Awareness.}
Interaction risk is determined through semantic cues such as the presence of lead vehicles and pedestrians. We define an \textit{unsafe-following ratio} to measure whether the ego vehicle maintains insufficient headway relative to its speed. Conversely, we also define a \textit{safe-following ratio} to capture especially cautious behavior. These quantities are computed over time and used to modulate style inference.

\paragraph{Rule\&Distribution-based Classification.}
Driving style is assigned by integrating maneuver type, interaction context, and temporal motion features:

\begin{itemize}[leftmargin=1em, topsep=0pt, itemsep=0.3em]
    \item \textbf{Aggressive (A):} Inferred when the ego vehicle exhibits insufficient headway to a lead agent, high average speed, or rapid acceleration profiles. This is particularly relevant during turns or when pedestrians are present, where aggressive dynamics raise potential safety concerns.
    
    \item \textbf{Conservative (C):} Assigned when the ego maintains large margins to other agents, exhibits smooth and slow motion, or maintains consistently safe gaps. Acceleration trends with minimal variability and substantial following distance also suggest conservative intent.
    
    \item \textbf{Normal (N):} Assigned in cases where the behavior does not strongly deviate toward either aggressiveness or caution, reflecting balanced dynamics.
\end{itemize}

\paragraph{Scene-Type Modulation.}
Though the classification rule is unified, its application is modulated by interaction context:

\begin{itemize}[leftmargin=1em, topsep=0pt, itemsep=0.3em]
    \item \textbf{Turning Maneuvers (Left/Right):} Risk is elevated due to lateral trajectory change and frequent presence of conflicting traffic. Aggressive classification emerges when the vehicle shows assertive dynamics under close headway; conservative behavior arises from cautious yielding or wide gaps.
    
    \item \textbf{Straight Crossings:} Style differentiation is informed by motion trends—e.g., acceleration, deceleration, or constant speed—combined with lead proximity.
    
    \item \textbf{Control Influence:} Protected intersections (e.g., traffic lights) allow slightly more assertive behavior without penalty, while unprotected junctions impose stricter expectations for caution.
\end{itemize}

\subsection{Heuristics in Crosswalk Scenarios}

Crosswalk scenarios require heightened awareness due to potential pedestrian interaction, despite typically lacking complex traffic geometry or leading vehicles. The classification of driving style in this context relies solely on ego-vehicle motion characteristics and the presence of pedestrians.

\paragraph{Contextual Awareness.}
Each scene includes a binary indicator \texttt{pedestrians} specifying whether pedestrians are present near the crosswalk area. No lead vehicle information is considered, and the road is assumed to be straight by default.

\paragraph{Motion-Based Features.}
We extract key kinematic features:
\begin{itemize}[leftmargin=1em, topsep=0pt, itemsep=0.3em]
    \item Average velocity $v_{\text{avg}}$;
    \item Maximum acceleration $a_{\max}$;
    \item Acceleration standard deviation $\sigma_a$;
    \item Motion trend (i.e., \textit{accelerating}), detected using linear regression over the velocity sequence.
\end{itemize}

\paragraph{Rule\&Distribution-based Classification.}
Style labels are assigned as follows:
\begin{itemize}[leftmargin=1em, topsep=0pt, itemsep=0.3em]
    \item \textbf{Aggressive (A):} If any of the following holds:
    \begin{itemize}
        \item $v_{\text{avg}} > \theta_v$;
        \item $a_{\max} > \theta_a$;
        \item $\sigma_a > \theta_\sigma$;
        \item motion trend indicates acceleration.
    \end{itemize}
    where $\theta_v$, $\theta_a$, and $\theta_\sigma$ are empirically chosen thresholds.
    
    \item \textbf{Conservative (C):} If $v_{\text{avg}}$ is significantly low (e.g., $< 2.0$~m/s), regardless of other indicators.
    
    \item \textbf{Normal (N):} All other cases.
\end{itemize}

\subsection{Heuristics in Side-to-Main (Ego on Main Road) Scenarios}

In \textit{side-to-main} scenarios—where the ego vehicle attempts to merge from a side road onto a main road—driving style is primarily influenced by gap acceptance behavior and merging aggressiveness. These scenarios demand real-time decision-making to balance caution and initiative, particularly under merging pressure.

\paragraph{Merging Context Identification.}
Scenes are annotated with a binary indicator \texttt{has\_merging}, which denotes whether main-road vehicles are present at the time of ego’s merging. This context shapes the interpretation of speed and acceleration—aggressive merging under pressure may reflect insufficient gap assessment.

\paragraph{Motion Features.}
We use the following kinematic attributes to infer driving style:
\begin{itemize}[leftmargin=1em, topsep=0pt, itemsep=0.3em]
    \item Average speed $v_{\text{avg}}$;
    \item Maximum total acceleration $a_{\max}$;
    \item Longitudinal velocity trend (\textit{accelerating}), estimated via linear regression.
\end{itemize}

\paragraph{Rule\&Distribution-based Classification.}
Classification follows this logic:
\begin{itemize}[leftmargin=1em, topsep=0pt, itemsep=0.3em]
    \item \textbf{Aggressive (A):}
    \begin{itemize}
        \item If \texttt{has\_merging} is \texttt{True} and $v_{\text{avg}}$ or $a_{\max}$ is high, or the velocity trend indicates acceleration;
        \item If \texttt{has\_merging} is \texttt{False} but $v_{\text{avg}}$ or $a_{\max}$ is high, indicating assertive motion despite absence of merging pressure.
    \end{itemize}
    
    \item \textbf{Conservative (C):} Assigned if $v_{\text{avg}}$ is low, suggesting a yielding or slow merge, regardless of merging context.

    \item \textbf{Normal (N):} All remaining cases with moderate dynamics and no acceleration spikes.
\end{itemize}

\subsection{Heuristics in Side-to-Main (Ego on Side Road) Scenarios}

In \textit{side-to-main} scenarios—where the ego vehicle is on the main road and may encounter merging traffic from a side road—driving style reflects how assertively or cautiously the ego navigates potential merging conflicts. This is especially relevant when assessing risk-aversion or dominance over right-of-way.

\paragraph{Interaction Context.}
The key semantic indicator \texttt{main\_road\_vehicles} signals whether merging traffic is present. Though pedestrian presence may also exist, classification primarily focuses on ego vehicle dynamics in response to merging activity.

\paragraph{Key Features.}
We use the following features to infer behavior:
\begin{itemize}[leftmargin=1em, topsep=0pt, itemsep=0.3em]
    \item Average velocity $v_{\text{avg}}$;
    \item Maximum acceleration $a_{\max}$;
    \item Acceleration standard deviation $\sigma_a$.
\end{itemize}

\paragraph{Rule\&Distribution-based Classification.}
\begin{itemize}[leftmargin=1em, topsep=0pt, itemsep=0.3em]
    \item \textbf{Aggressive (A):} Assigned if both average speed and peak acceleration are high, i.e., the ego maintains assertive forward motion under merging conditions or in general. This includes patterns such as:
    \[
    v_{\text{avg}} > \tau_v^{A}, \quad a_{\max} > \tau_a^{A}
    \]
    where $\tau_v^{A}$ and $\tau_a^{A}$ are empirically set thresholds.

    \item \textbf{Conservative (C):} Assigned if the ego vehicle exhibits consistently low average speed and stable acceleration, indicating cautious navigation:
    \[
    v_{\text{avg}} < \tau_v^{C}, \quad \sigma_a < \tau_\sigma^{C}
    \]

    \item \textbf{Normal (N):} All cases in between, with moderate speed and neither aggressive nor cautious acceleration profiles.
\end{itemize}

\subsection{Heuristics in Lane Change Scenarios}

Lane change scenarios involve dynamic and potentially high-risk lateral maneuvers, requiring the ego vehicle to coordinate its motion with adjacent traffic. Driving style in this context is determined by the intended direction (left or right), spatial constraints (e.g., rear or front vehicle presence), and motion characteristics such as lateral velocity and acceleration variability.

\paragraph{Direction and Rear Vehicle Awareness.}
The intended lane change direction is inferred from yaw change $\Delta \psi$: left lane changes are detected when $\Delta \psi > 0.25$ radians, and right lane changes when $\Delta \psi < -0.25$. Rear vehicle presence in the target lane is denoted by scene indicators \texttt{has\_left\_rear} and \texttt{has\_right\_rear}. Additional indicators such as front vehicle proximity are used to determine urgency-driven maneuvers.

\paragraph{Motion Indicators.}
The following dynamic features are extracted to characterize maneuver intensity and stability:
\begin{itemize}[leftmargin=1em, topsep=0pt, itemsep=0.3em]
    \item Maximum absolute lateral velocity: $v_y^{\max} = \max |v_y|$;
    \item Average longitudinal velocity: $v_{\text{avg}}$;
    \item Maximum acceleration: $a_{\max}$;
    \item Acceleration standard deviation: $\sigma_a$;
    \item Yaw difference during maneuver: $\Delta \psi$.
\end{itemize}

\paragraph{Rule\&Distribution-based Classification.}
Rules vary depending on lane-change direction and whether rear vehicles are present:

\begin{itemize}[leftmargin=1em, topsep=0pt, itemsep=0.3em]
    \item \textbf{Aggressive (A):}
    \begin{itemize}
        \item \textit{Left change with rear vehicle:} If any of $a_{\max}$, $\sigma_a$, or yaw change $\Delta \psi$ exceeds threshold—regardless of speed—indicating abrupt motion under risky context;
        \item \textit{Left change without rear vehicle:} Only if both $v_{\text{avg}}$ is high and at least one of $a_{\max}$, $\sigma_a$, or $\Delta \psi$ is large;
        \item \textit{Right change with rear vehicle:} Similar to left—if $a_{\max}$, $\sigma_a$, or $\Delta \psi$ is high, label as aggressive;
        \item \textit{Right change without rear vehicle:} Must satisfy $v_{\text{avg}}$ being high \textbf{and} any of $a_{\max}$, $\sigma_a$, or $\Delta \psi$ being high.
    \end{itemize}
    
    \item \textbf{Conservative (C):}
    \begin{itemize}
        \item \textit{Left change:} When $\Delta \psi$ and $\sigma_a$ are both low, indicating smooth direction change;
        \item \textit{Right change:} When $\Delta \psi$ and $v_{\text{avg}}$ are both low, reflecting gentle, controlled motion.
    \end{itemize}

    \item \textbf{Normal (N):} Default for samples not matching extreme behaviors.
\end{itemize}

\subsection{Heuristics in Countryside Road Scenarios}

Countryside road scenarios are characterized by low traffic density and relaxed speed regulations, yet may involve substantial geometric variability—such as extended straightaways and sharp curves. These conditions afford greater behavioral latitude, necessitating classification rules that distinguish between permissible cruising and potentially hazardous or overly passive maneuvers.

\paragraph{Geometric Classification.}
Each scene is annotated as either \texttt{Straight} or \texttt{Curve}, based on map topology and trajectory shape. This binary classification contextualizes motion interpretation, as curved roads generally require tighter control and greater caution.

\paragraph{Motion Features.}
We extract the following indicators of ego dynamics to infer driving style:
\begin{itemize}[leftmargin=1em, topsep=0pt, itemsep=0.3em]
    \item Average velocity $v_{\text{avg}}$;
    \item Maximum acceleration $a_{\max}$;
    \item Acceleration standard deviation $\sigma_a$.
\end{itemize}

\paragraph{Rule\&Distribution-based Classification.}
Rules are unified across straight and curved countryside settings, with style assigned as follows:

\begin{itemize}[leftmargin=1em, topsep=0pt, itemsep=0.3em]
    \item \textbf{Aggressive (A):} 
    Labeled if \textbf{any} of the motion indicators—$v_{\text{avg}}$, $a_{\max}$, or $\sigma_a$—exceed predefined thresholds, indicating assertive or potentially unstable behavior regardless of road geometry.

    \item \textbf{Conservative (C):}
    Assigned if \textbf{both} $v_{\text{avg}}$ and $\sigma_a$ are low, reflecting slow and smooth progression consistent with cautious behavior.

    \item \textbf{Normal (N):}
    Default label for cases exhibiting moderate speed and variability, falling between the two extremes.
\end{itemize}

\subsection{Heuristics in Roundabout Entrance Scenarios}

Roundabout entrance scenarios involve complex maneuvering under potential merging pressure. Driving style is categorized into \textit{Aggressive (A)}, \textit{Normal (N)}, or \textit{Conservative (C)} based on ego-vehicle velocity patterns, acceleration behavior, and contextual merging risk. A rule-based strategy is used to ensure interpretability and context sensitivity.

\paragraph{Contextual Modulation.}
Each scene is annotated with a binary indicator \texttt{merge\_risk}, signaling whether surrounding vehicles are likely to merge into the roundabout concurrently. This flag modulates the aggressiveness threshold: in the presence of merging risk, even moderately assertive behaviors may be treated as aggressive, while cautious entries are emphasized as conservative.

\paragraph{Dynamic Features.}
The classification is based on a combination of behavioral indicators, including:
\begin{itemize}[leftmargin=1em, topsep=0pt, itemsep=0.3em]
    \item Ego vehicle's average speed and longitudinal acceleration pattern;
    \item Variability in acceleration, as a proxy for control smoothness;
    \item Velocity trends derived from temporal fitting (e.g., accelerating, decelerating, quasi-constant).
\end{itemize}

\paragraph{Rule\&Distribution-based Classification.}
The driving style is assigned through the joint evaluation of motion dynamics and contextual semantics:

\begin{itemize}[leftmargin=1em, topsep=0pt, itemsep=0.3em]
    \item \textbf{With merge risk:}
    \begin{itemize}
        \item \textit{Aggressive (A):} Identified when the ego vehicle exhibits assertive acceleration or non-monotonic trends associated with fast entry;
        \item \textit{Conservative (C):} Assigned if the vehicle maintains slow, steady motion or shows clear deceleration before merging;
        \item Remaining behaviors are labeled as \textbf{Normal (N)}.
    \end{itemize}

    \item \textbf{Without merge risk:}
    \begin{itemize}
        \item \textit{Aggressive (A):} Assigned for high-speed or strong-acceleration entries, even in low-risk settings;
        \item \textit{Conservative (C):} Assigned for low-speed and smooth trajectories;
        \item Intermediate behaviors default to \textbf{Normal (N)}.
    \end{itemize}
\end{itemize}

\section{Training Details of Benchmark Methods and VLM Fine-tuning}\label{sec: vlm}

\subsection{Details of Benchmark Methods Training}
All benchmark models were trained on a Linux server equipped with 8× NVIDIA A100 GPUs and dual AMD EPYC 7742 64-core CPUs. Each individual model was trained using two A100 GPUs, with a batch size fixed at 64. For models incorporating style-conditioning, the number of training epochs was matched to that of their non-style-conditioned counterparts and was set equal to the number of epochs used by the original naive baseline, ensuring a fair comparison across all settings.

Regarding optimization, the learning rate for both DiffusionDrive~\cite{liao2024diffusiondrive} and DiffusionDrive-Style was set to 6e-4. For AD-MLP\cite{zhai2023rethinking}, Transfuser~\cite{chitta2022transfuser}, WoTE~\cite{li2025end}, as well as their style-conditioned variants (AD-MLP-Style, Transfuser-Style, WoTE-Style), the learning rate was consistently set to 1e-4.

\subsection{Details of VLM Fine-tuning towards Better Driving Scene Understanding}
To enable accurate and context-sensitive semantic reasoning in driving scenes, we fine-tune the Video-LLaMA3 model~\cite{zhang2025videollama} using lightweight parameter-efficient tuning (LoRA). This process tailors the model to better understand autonomous driving scenarios, which are underrepresented in generic vision-language pretraining.

\paragraph{Training Data.}
We build upon the LingoQA dataset~\cite{marcu2024lingoqa}, a multimodal benchmark comprising video clips and scene-level question-answer pairs focused on road semantics, interactions, and intent recognition. The dataset includes approximately 267.8k video–prompt–response triplets, covering diverse topology and behavioral cases.

\paragraph{Model Setup.}
We adopt Video-LLaMA3 as the base model due to its strong multimodal reasoning capacity and open-source support. For fine-tuning:
\begin{itemize}[leftmargin=1em, topsep=0pt, itemsep=0.3em]
    \item We freeze the vision encoder and most transformer layers.
    \item We insert LoRA adapters in the query and value projection matrices of selected attention layers.
    \item We train for 3 epochs on 6×A100 GPUs, a learning rate of 1e-4, and batch size of 96.
    \item The fine-tuning was completed in under 8 GPU-hours, making it suitable for scalable deployment.
\end{itemize}

\end{document}